%% file: main.tex
\documentclass{article}

\usepackage{conference,times}
\input{math_commands.tex}

\usepackage[utf8]{inputenc}
\usepackage[T1]{fontenc}
\usepackage{amsmath,amssymb,amsfonts}
\usepackage{booktabs}
\usepackage{wrapfig}
\usepackage{graphicx}
\usepackage{tikz}
\usepackage{fancyhdr}
\usetikzlibrary{arrows.meta,positioning,shapes.geometric}
\usepackage{enumitem}
\usepackage{listings}
\usepackage{microtype}
\usepackage{pifont}
\usepackage[table]{xcolor}
\usepackage{multirow}
\usepackage{makecell}
\usepackage{placeins}
\usepackage[most]{tcolorbox}

\definecolor{iccvblue}{rgb}{0.21,0.49,0.74}

\usepackage[
    breaklinks=true,
    colorlinks=true,
    citecolor=iccvblue,
    urlcolor=iccvblue,
    linkcolor=black
]{hyperref}
\usepackage{url}
  
  \definecolor{promptbg}{RGB}{248,249,251}
  \definecolor{promptframe}{RGB}{210,214,220}
  \definecolor{prompttitle}{RGB}{35,45,60}

  \lstdefinestyle{promptstyle}{
      basicstyle=\ttfamily\scriptsize,
      breaklines=true,
      breakatwhitespace=false,
      columns=fullflexible,
      keepspaces=true,
      showstringspaces=false,
      frame=none,
      xleftmargin=0pt,
      xrightmargin=0pt,
      aboveskip=2pt,
      belowskip=2pt
  }

  \newtcolorbox{promptbox}[1]{
      enhanced,
      breakable,
      colback=promptbg,
      colframe=promptframe,
      coltitle=prompttitle,
      fonttitle=\bfseries\small,
      title=#1,
      boxrule=0.5pt,
      arc=2pt,
      left=6pt,
      right=6pt,
      top=5pt,
      bottom=5pt,
      before skip=6pt,
      after skip=8pt
  }

\definecolor{best}{HTML}{E8F5E9}
\definecolor{best2}{HTML}{E3F2FD}

\title{IndusAgent: Agentic Reinforcement Learning for Cross-Category Industrial Anomaly Detection}

\author{%
\makebox[\textwidth][c]{%
\begin{minipage}[t]{1.1\textwidth}
\centering
\normalfont\normalsize
\textbf{Rongbin Tan}$^{1,2,*}$ \quad
\textbf{Fangfang Lin}$^{3,*}$ \quad
\textbf{Jinguo Luo}$^{4,*}$ \quad
\textbf{Min Qiu}$^{5}$ \\[5pt]
\textbf{Kejin Cui}$^{6}$ \quad
\textbf{Mengmeng Wang}$^{7}$ \quad
\textbf{Yi Wang}$^{1}$ \quad
\textbf{Jiyuan Wang}$^{8}$ \\[5pt]
\textbf{Yue Wang}$^{9,\ddagger}$ \quad
\textbf{Zhenlong Yuan}$^{6}$ \quad
\textbf{Shuhan Song}$^{1,\S}$ \quad
\textbf{Huawei Cao}$^{1,2}$ \\[10pt]
$^{1}$~State Key Laboratory of Processors \quad
$^{2}$~Nanjing College, University of Chinese Academy of Sciences \\[5pt]
$^{3}$~Santa Clara University \quad
$^{4}$~Kling Team, Kuaishou Technology  \quad
$^{5}$~Trine University \\[5pt]
$^{6}$~Independent Researcher \quad
$^{7}$~New York University \quad
$^{8}$~Beijing Jiaotong University \quad
$^{9}$~Meta Platforms \\[10pt]
{\small\itshape
$^{*}$~Equal contribution \qquad
$^{\ddagger}$~Project Lead \qquad
$^{\S}$~Corresponding Author
}
\end{minipage}%
}%
}

\iclrfinalcopy 

\begin{document}

\maketitle
\lhead{} 

\begin{abstract}
Small defects in unfamiliar product categories can be difficult to distinguish from benign appearance variations in a global view. Additional reasoning over the same view may leave this ambiguity unresolved. We introduce \textbf{IndusAgent}, a Qwen3-VL-8B inspection policy trained through supervised fine-tuning and tool-augmented reinforcement learning for cross-category industrial anomaly detection. The policy selects among six tools for cropping, enhancement, measurement, within-image comparison, texture analysis, and external visual opinions from Qwen3-VL-Max, then integrates returned observations into a structured diagnosis. Indus-CoT provides 3,000 trajectories from 26 Real-IAD categories, generated with actual tool feedback and screened against annotations after teacher generation. GRPO optimizes an accuracy-gated objective that conditions localization, type, and tool rewards, including call costs, on final diagnostic correctness while retaining an independent format reward. Across five benchmarks, IndusAgent achieves 83.5\% mean balanced accuracy, compared with 79.4\% for direct Max inference under the common protocol. Across three training seeds, the five-benchmark mean balanced accuracy is 80.5\% for a weight-matched ungated control and 83.5\% for the gated policy. Evaluation uses disjoint product categories and a fixed defect vocabulary. 

\end{abstract}

\section{Introduction}
Industrial anomaly detection identifies defective products and localizes the visual evidence supporting an inspection decision~\citep{mvtec,visa}. Reconstruction and feature-based methods learn normal appearance from product-specific examples~\citep{zavrtanik2021draem,roth2022patchcore}. When the inspected product changes, acquiring representative normal images can become a practical constraint. We study a category-disjoint, reference-free setting in which target product categories are excluded from industrial post-training and no paired normal exemplar is provided. 

\begin{figure*}[t]
\centering
\includegraphics[width=\textwidth]{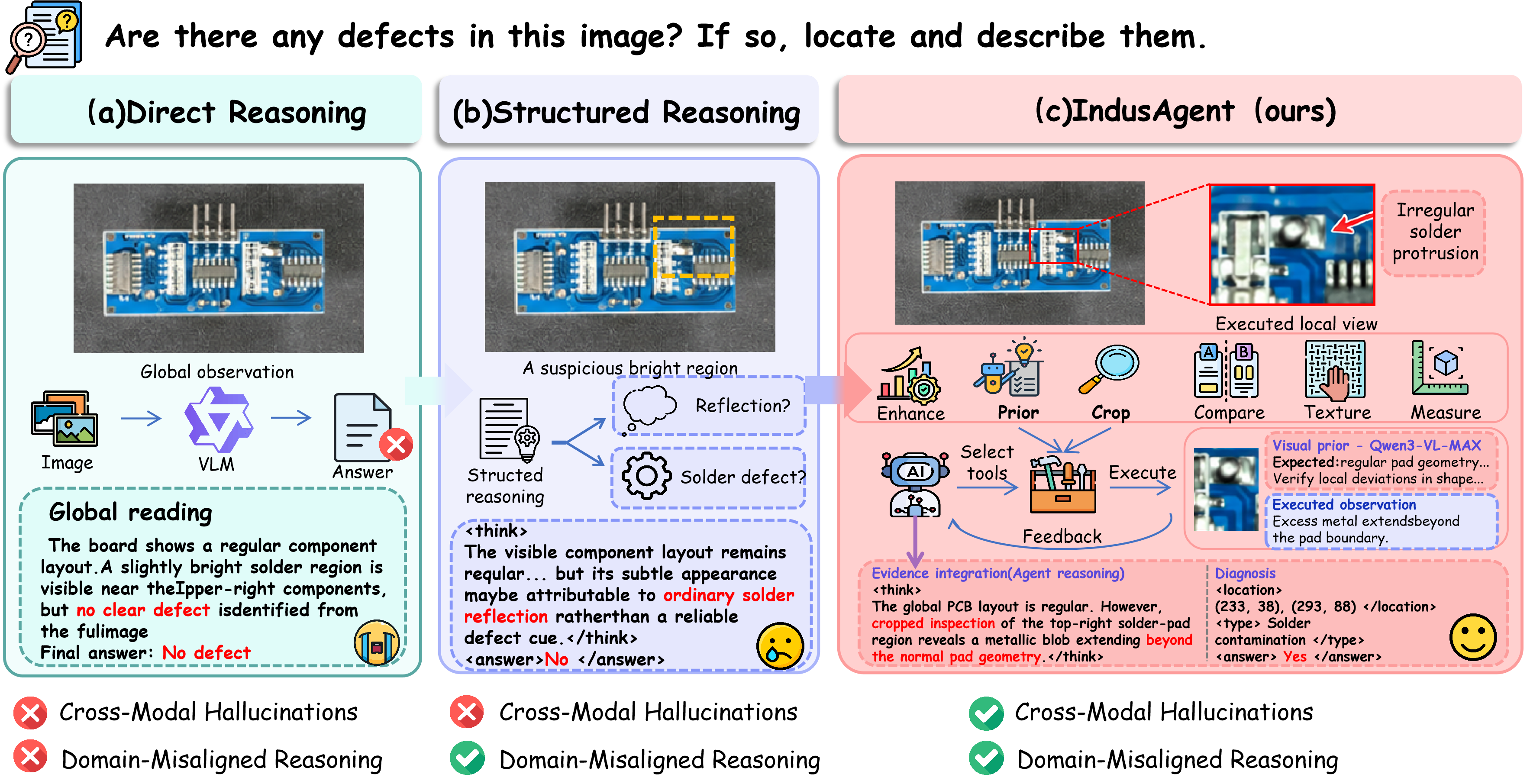}
\caption{\textbf{Inspection beyond the global view.} On the illustrated PCB query, direct reasoning and structured reasoning produce a no-defect decision, while IndusAgent inspects a returned local crop and incorporates an external visual opinion before predicting an anomaly. The three columns illustrate alternative inspection paths for this example. The visual-prior tool queries Qwen3-VL-Max.}
\label{fig:intro}
\end{figure*}

Vision-language models support inspection through text-image alignment and visual reasoning~\citep{jeong2023winclip,gu2024anomalygpt,xu2025towards}. Multimodal backbones provide the visual-language interface~\citep{verifiedOpenAI2023GPT4V,Qwen2.5-VL,Qwen3-VL}, and industrial post-training adapts it to diagnostic outputs~\citep{li2026iadr1,chao2025anomalyr1}. Small defects can nevertheless remain ambiguous in a global view. In Figure~\ref{fig:intro}, the structured-reasoning path attributes a suspicious bright region to a possible reflection and retains a normal decision. IndusAgent instead requests additional observations and predicts an anomaly after local inspection. This example motivates learning how to acquire and interpret further evidence when the initial view leaves a diagnosis uncertain.

Tool-assisted inspection extends the evidence available beyond a global assessment. AgentIAD combines tool-integrated trajectories, supervised fine-tuning, and reinforcement learning, and includes both local-only inspection and normal-exemplar retrieval modes~\citep{miao2025agentiad}. We study how to learn inspection actions from teacher-selected calls and executed feedback under product-category exclusion. Ground-truth annotations screen completed teacher trajectories rather than determine their inspection actions. During policy optimization, final diagnostic correctness determines whether auxiliary inspection outputs receive credit. These choices connect trajectory construction and reward design to the cross-category inspection setting.

We introduce \textbf{IndusAgent}, a tool-augmented reinforcement learning framework for cross-category industrial anomaly detection. Its Qwen3-VL-8B policy selects among six tools for local cropping, visual enhancement, geometric measurement, within-image comparison, texture analysis, and an external visual prior. Comparison uses regions from the same query image. The prior queries Qwen3-VL-Max with the input image for expected normal structure and visible deviations requiring verification. The resulting system combines an 8B policy with an external API and integrates the returned images and text into a structured prediction.

Training begins with \emph{Indus-CoT}, a corpus of 3,000 trajectories from 26 Real-IAD categories. The teacher receives the query image, selects tools, and integrates actual execution feedback without seeing ground-truth annotations. Completed trajectories are filtered against annotations to initialize the policy through supervised fine-tuning. Online GRPO then optimizes the two-round inspection trajectories. Training and inference both use two responses. When no tool is selected, the call and observation lists are empty, and the second response still provides the final diagnosis.

Our reward design makes final diagnostic correctness a prerequisite for auxiliary inspection credit. An additive objective can reward an incorrect diagnosis when the rollout nevertheless contains annotation-matching auxiliary fields or a tool-related gain. The accuracy gate removes such credit while retaining a separate format reward to supervise the response structure. Tool-call costs share this gate: they penalize additional calls on correctly diagnosed trajectories, while an incorrect trajectory receives only its format reward. We evaluate this coupling against an additive objective with the same reward terms and coefficients. Our contributions are:
\begin{itemize}
    \item \textbf{Tool-Augmented Industrial Inspection.}
    We propose \textbf{IndusAgent}, a framework that enables a Qwen3-VL-8B policy to select among six tools and integrate their feedback for cross-category industrial anomaly detection. It combines local inspection with an external visual prior, requiring no paired normal exemplar at inference.

    \item \textbf{Indus-CoT Dataset and Agentic Training.}
    We construct \textbf{Indus-CoT}, a dataset of 3,000 inspection trajectories from 26 Real-IAD categories, generated with teacher-selected tool calls, executed feedback, and post-generation annotation screening. Building on this dataset, we develop a two-stage SFT--GRPO training pipeline with an accuracy-gated objective that conditions auxiliary inspection rewards and call costs on final diagnostic correctness while supervising format independently.

    \item \textbf{Empirical Effectiveness.}
    IndusAgent achieves 83.5\% mean balanced accuracy across five industrial anomaly detection benchmarks, exceeding direct Qwen3-VL-Max inference by 4.1 percentage points under the common evaluation protocol. Across three training seeds, the accuracy-gated objective further yields a 3.0-point mean improvement over a weight-matched ungated objective.
\end{itemize}

\section{Methodology}

\begin{figure} [!t]
  \centering
  \includegraphics[width=\linewidth]{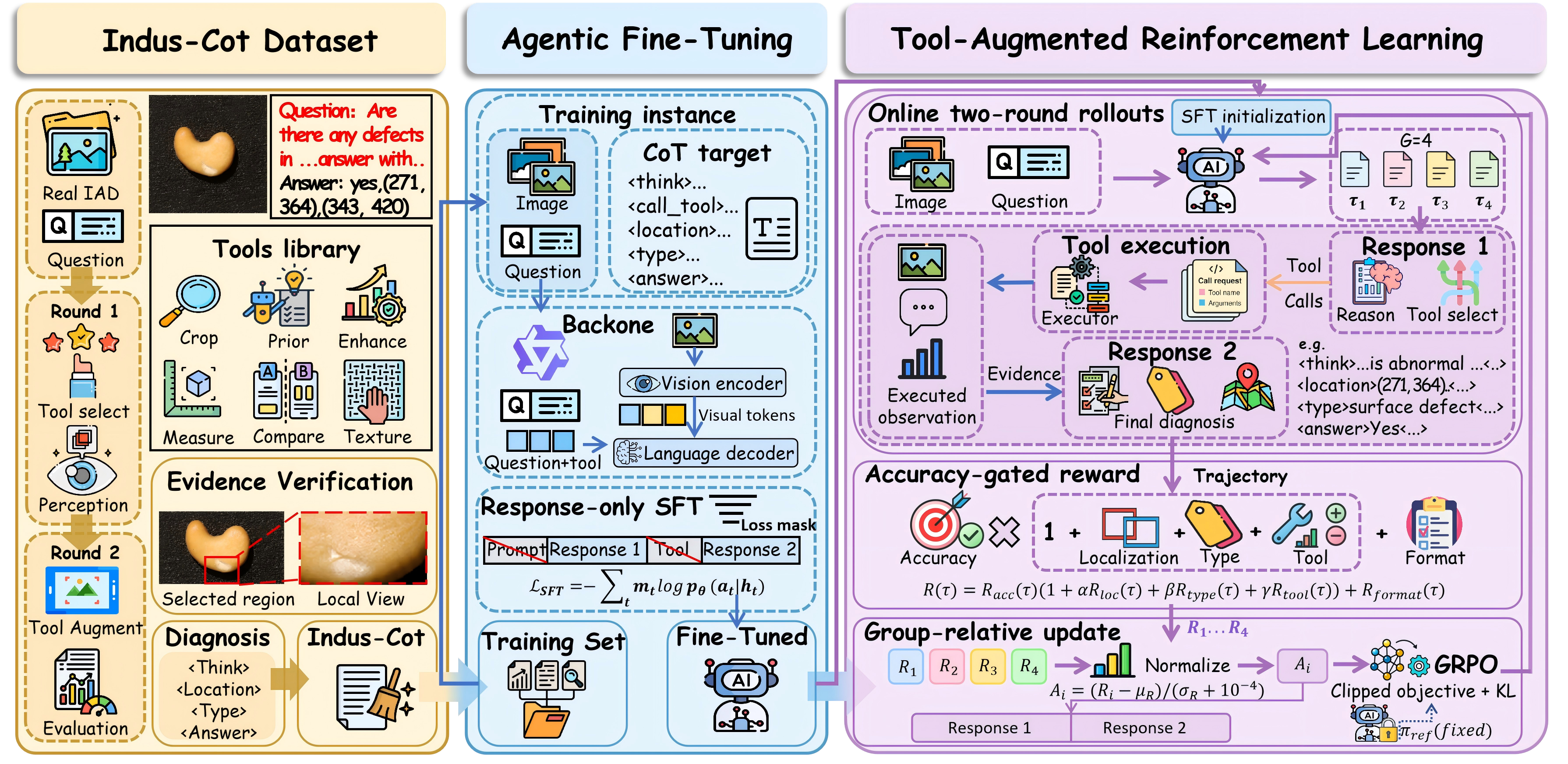}
  \caption{\textbf{Overview of the IndusAgent pipeline.}
(i) Indus-CoT contains 3,000 screened trajectories from 26 Real-IAD categories, pairing teacher-selected tool calls with executed feedback.
(ii) Response-only SFT initializes the inspection policy.
(iii) Online two-round GRPO refines the policy with an accuracy-gated reward, conditioning auxiliary inspection rewards and call costs on final diagnostic correctness while supervising format independently.}
  \label{fig:overview1}
\end{figure}

\textbf{Overview.}
IndusAgent consists of teacher-trajectory construction, supervised policy initialization, and online tool-augmented GRPO (Fig.~\ref{fig:overview1}). We first screen generated trajectories against source annotations, then train the policy to produce structured responses, and finally optimize the policy with the accuracy-gated reward.

\subsection{Systematic Definition and Agentic Toolkit}
\label{sec:Formulation}
\textbf{Problem Formulation.}
Given a query image $I\in\mathbb{R}^{H\times W\times3}$ and instruction $Q$, the policy $\pi_\theta$ produces a binary diagnosis and an inspection trace. When predicting an anomaly, it also returns a final defect box and a label from a fixed canonical defect vocabulary. We represent the final defect box as $B=[x_1,y_1,x_2,y_2]$
in original-image pixel coordinates, with $(x_1,y_1)$ and
$(x_2,y_2)$ denoting the top-left and bottom-right corners,
respectively. The backbone is Qwen3-VL-8B-Instruct~\citep{Qwen3-VL}. Source training categories are disjoint from the evaluation categories, and no paired normal exemplar is supplied at inference. 

\textbf{Fixed Two-round Inference with Optional Tool Feedback.}
Training and inference use the same two-round protocol.
The first response produces a preliminary diagnosis and selects zero or more calls from
$\mathcal T=\{T_{\mathrm{crop}},T_{\mathrm{prior}},T_{\mathrm{enhance}},T_{\mathrm{measure}},T_{\mathrm{compare}},T_{\mathrm{texture}}\}$.
The environment executes the selected calls and returns their observations.
If no tool is selected, the call and observation lists are both empty.
A second response is generated in every case:
\begin{equation}
a^{(1)}\sim\pi_\theta(\cdot\mid I,Q),\qquad
a^{(2)}\sim\pi_\theta(\cdot\mid I,Q,a^{(1)},o),
\end{equation}
where $o$ is the observation list.
Only the second response provides the final diagnosis and, for an anomalous prediction, the defect box in original-image pixel coordinates and canonical type.
Appendix~\ref{appx:prompt} specifies the shared output schema.

\begin{figure}[t]
    \centering
    \includegraphics[width=\linewidth]{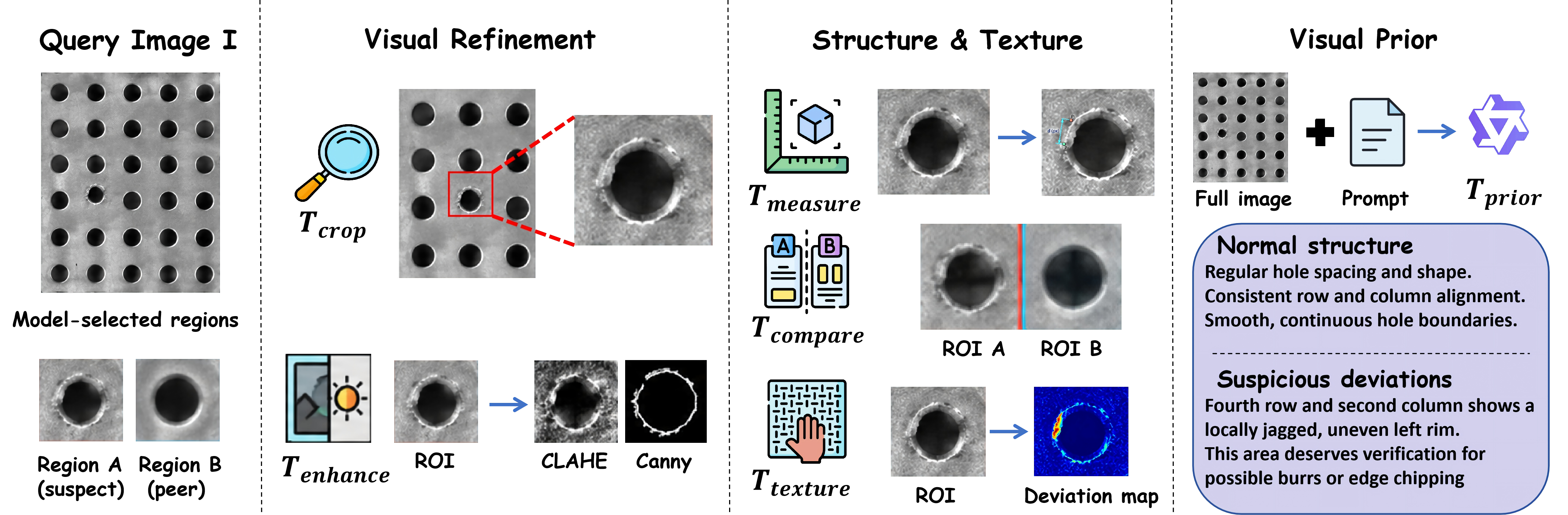}
    \caption{\textbf{Six-tool inspection library.} Crop, enhancement, measurement, within-image comparison, and texture analysis provide observations derived from the query image. The comparison regions shown here come from the same query, without a paired normal exemplar. The visual-prior tool queries the external Qwen3-VL-Max API for descriptions of structure and visible deviations. }

    \label{fig:tool}
\end{figure}

\textbf{Agentic Toolkit.} As shown in Fig.~\ref{fig:tool}, we equip IndusAgent with six complementary tools that provide visual, quantitative, and textual observations for category-disjoint industrial anomaly detection. All six are available during training and target inference.

\textbf{(1)~Crop} ($T_{\mathrm{crop}}$) extracts model-selected regions;
\textbf{(2)~Enhance} ($T_{\mathrm{enhance}}$) applies CLAHE and Canny operators to reveal local details;
\textbf{(3)~Measure} ($T_{\mathrm{measure}}$) computes pixel distances and angles;
\textbf{(4)~Compare} ($T_{\mathrm{compare}}$) contrasts similar regions within the same query image;
\textbf{(5)~Texture} ($T_{\mathrm{texture}}$) characterizes local texture variations;
\textbf{(6)~Visual Prior} ($T_{\mathrm{prior}}$) queries Qwen3-VL-Max with the full query image to describe expected normal structure and suspicious deviations, without returning a final Yes/No diagnosis.

\subsection{Indus-CoT Dataset}
\label{sec:Training}

We construct \textbf{Indus-CoT} for supervised initialization using 3,000 Real-IAD trajectories, balanced between 1,500 normal and 1,500 anomalous images~\citep{wang2024realiad}. Four product categories overlapping with the evaluation benchmarks are excluded before construction, leaving 26 categories.

\begin{itemize}[leftmargin=*,nosep]
\item \textbf{Inspection and tool selection.}
Given the query image and task instruction, the teacher forms a preliminary diagnosis and selects tools and arguments without ground-truth labels, boxes, types, or a paired normal reference.

\item \textbf{Tool execution and diagnosis.}
The environment executes the selected calls and returns the resulting images or text. The teacher integrates this feedback into a final diagnosis and, when predicting an anomaly, a defect box and type.

\item \textbf{Post-generation screening.}
Completed trajectories are checked against source annotations for output format, binary correctness, and required auxiliary fields. Anomalous samples must match the canonical defect type and achieve box IoU $\geq 0.5$. Normal samples must end with \texttt{No} and no final box or type.
\end{itemize}

Initial screening retains 2,747 trajectories. We obtain 253 accepted replacements from additional images in the same source categories using the same procedure. Failed candidates are discarded without annotation-guided repair. Appendix~\ref{appx:exp_details} provides the category audit and screening details.

\subsection{Supervised Fine-Tuning}
\label{sec:SFT}

We initialize Qwen3-VL-8B-Instruct with supervised fine-tuning on Indus-CoT before online reinforcement learning. This stage provides supervision for the diagnostic output schema, tool routing, and interpretation of returned evidence. The no-SFT ablation evaluates the effect of this initialization within our training setup; it does not establish that SFT is universally necessary for inspection.

Formally, let $A=(a_1,\ldots,a_T)$ contain all policy-generated response tokens, including reasoning, tool routing and arguments, the final binary answer, and the optional location and type fields. Prompt tokens and externally returned tool observations are provided as context but are masked from the loss. With $m_t\in\{0,1\}$ denoting this response mask, SFT minimizes
\begin{equation}
\mathcal{L}_{\text{SFT}} = -\mathbb{E}_{\tau \sim \mathcal{D}}
\left[ \sum_{t=1}^{T} m_t \log \pi_{\theta}(a_t \mid h_t) \right].
\end{equation}
Here $h_t$ includes the query, instruction, preceding responses, and any executed observations available before token $t$. The objective supervises the policy output, including the text inside \texttt{<think>}; environment-returned observations remain non-differentiable context for the next response.

\subsection{Agentic Reinforcement Learning}
\label{sec:RL}

\textbf{Online Two-round Rollouts.}
For each source query, the rollout policy $\pi_{\theta_{\mathrm{old}}}$ samples $G$ first responses. Each response selects zero or more calls; the environment executes them and returns the observations. A second response is generated for every trajectory, including no-tool trajectories. Ground-truth labels, boxes, and types are supplied to reward computation, not to the policy input. Both policy-generated response segments contribute to the loss; prompts and environment-returned observations do not. Inference follows the same two-round protocol described in Section~\ref{sec:Formulation}.

\textbf{Group Relative Policy Optimization (GRPO).}
We adopt GRPO~\citep{shao2024deepseekmath} for policy optimization. For each query $q$, we sample $G$ rollout trajectories $\{o_1,\ldots,o_G\}$ from the previous policy $\pi_{\theta_{\mathrm{old}}}$ and compute their advantages from within-query reward comparisons.

The objective below uses compact trajectory-indexed notation. Within each trajectory contribution, probability ratios are evaluated separately at each policy-generated token, and clipping precedes token averaging. We first average the token losses within each trajectory and then average over sampled trajectories. Token indices and conditioning on the preceding history are suppressed: the ratios in the clipped objective refer to token-level conditional probabilities rather than joint trajectory probabilities. Both policy-generated response segments contribute to the loss; prompts and executed tool observations provide context only.
\begin{equation} 
\begin{aligned} 
    &\mathcal{L}_{GRPO}(\theta) = -\mathbb{E}_{q \sim P(Q), \{o_i\}_{i=1}^{G} \sim \pi_{\theta_{old}}(O|q)} \Bigg[ \frac{1}{G} \sum_{i=1}^G \\ 
    &\quad \Bigg( \min\left( \frac{\pi_\theta(o_i|q)}{\pi_{\theta_{old}}(o_i|q)} A_i, \text{clip}\left(\frac{\pi_\theta(o_i|q)}{\pi_{\theta_{old}}(o_i|q)}, 1-\epsilon, 1+\epsilon\right) A_i \right) - \beta \mathbb{D}_{KL}(\pi_\theta \| \pi_{ref}) \Bigg) \Bigg], 
\end{aligned} 
\end{equation} 
The KL expression below likewise suppresses token indices and history conditioning and denotes a sampled penalty relative to the fixed reference policy. KL computation is separate from groupwise reward normalization.
\begin{equation} 
\mathbb{D}_{KL}(\pi_\theta||\pi_{ref}) = \frac {\pi_{ref}(o_i|q)} {\pi_\theta(o_i|q)} - \log \frac {\pi_{ref}(o_i|q)} {\pi_\theta(o_i|q)} - 1, 
\end{equation} 
Here, $\epsilon$ controls clipping, and $\beta$ denotes the KL regularization coefficient within this subsection; it is distinct from the type-reward coefficient denoted by $\beta$ in the reward design. Each trajectory shares one advantage across its policy tokens. For a group with nonzero reward standard deviation, the advantage is
\begin{equation} 
A_{i} = \frac{r_i - \text{mean}(\{r_1, r_2, \dots, r_G\})}{\text{std}(\{r_1, r_2, \dots, r_G\})}. 
\end{equation} 
When all rewards in a group are equal, we set $A_i=0$ for every trajectory without evaluating the division above. Such a group contributes no reward-driven policy update. Each trajectory receives the composite reward defined below.

\textbf{Accuracy-gated Reward.}
For a completed trajectory $\tau$, final diagnostic correctness controls the contribution of the auxiliary inspection terms:
\begin{equation}
\label{eq:gated_reward}
R(\tau)=R_{\mathrm{acc}}(\tau)
\bigl(1+\alpha R_{\mathrm{loc}}(\tau)+\beta R_{\mathrm{type}}(\tau)+\gamma R_{\mathrm{tool}}(\tau)\bigr)
+R_{\mathrm{format}}(\tau),
\end{equation}
with $\alpha=0.8$, $\beta=0.6$, and $\gamma=0.5$. Binary correctness gives $R_{\mathrm{acc}}\in\{0,1\}$. For truly anomalous images, $R_{\mathrm{loc}}$ is continuous IoU between predicted and ground-truth boxes expressed in the same original-image pixel coordinate system, and $R_{\mathrm{type}}$ is exact agreement after deterministic canonicalization. Missing or invalid predictions receive zero, and both terms are zero for truly normal images. The teacher-screening threshold IoU$\geq0.5$ is separate from the continuous RL localization reward.

The tool term combines a confidence-gain bonus with a cost per parsed call:
\begin{equation}
\label{eq:tool_reward}
R_{\mathrm{tool}}=0.3\,\mathbb I[N_{\mathrm{ok}}>0\ \land\ \Delta m>0]-0.1N_{\mathrm{call}}.
\end{equation}
Here $N_{\mathrm{call}}$ counts up to four parsed calls admitted to execution, including repeated calls; excess requests are ignored. A counted call incurs cost even if execution fails. $N_{\mathrm{ok}}$ counts recognized observations with status \texttt{ok}. The margin change $\Delta m$ uses Yes/No logits oriented toward the final predicted label. It measures confidence change across the two response contexts, which also differ in reasoning history (Appendix~\ref{appx:reward_design}). A compliant two-round trajectory receives $R_{\mathrm{format}}=0.05$; otherwise this term is zero. With $N_{\mathrm{call}}\leq4$, correct diagnoses yield $R\geq0.8$, whereas incorrect diagnoses yield $R\leq0.05$.

\section{Experiments}

\subsection{Experimental Setup}
\label{sec:exp_setup}
\label{sec:experiments_setup}

\textbf{Datasets and Benchmarks.}
We evaluate \textbf{IndusAgent} on five industrial anomaly detection benchmarks: MVTec-AD~\citep{mvtec}, VisA~\citep{visa}, MPDD~\citep{verifiedJezek2021MPDD}, DTD~\citep{dtd}, and SDD~\citep{verifiedTabernik2019SDD}. 
These datasets cover two representative scenarios: (1) \textcolor{teal}{\emph{industrial objects}}, which are characterized by complex structures, poses, and geometries; (2) \textcolor{orange}{\emph{surface textures}}, where defects are often subtle and embedded within repetitive. 

\textbf{Common-protocol Comparisons.}
We evaluate the direct Qwen3-VL-8B backbone, direct Qwen3-VL-Max, IAD-R1, the full system, and the full system with the prior disabled at inference. These rows use the same images, preprocessing, binary task semantics, and metric definitions, with model-appropriate answer extraction described in Appendix~\ref{appx:baseline_parsing}.

\subsection{Main Results}

\begin{table}[htb]
\centering
\caption{Image-level balanced accuracy (\%) on five IAD benchmarks. Baseline results other than our Qwen3-VL-8B rerun are transcribed from IAD-R1~\citep{li2026iadr1}. IndusAgent combines an 8B policy with the Qwen3-VL-Max API.}
\label{model_comparison}
\resizebox{\textwidth}{!}{
\begin{tabular}{l|lc|ccc|cc|c}
\toprule
\multicolumn{1}{c|}{\multirow{2}{*}{Type}} &
\multicolumn{1}{c}{\multirow{2}{*}{Model}} &
\multicolumn{1}{c|}{\multirow{2}{*}{Param.}} &
\multicolumn{3}{c|}{\textbf{Industrial Workpieces}} &
\multicolumn{2}{c|}{\textbf{Surface Texture}} &
\multicolumn{1}{c}{\multirow{2}{*}{Avg.}} \\
\cmidrule(lr){4-6} \cmidrule(lr){7-8}
& & & MVTec & MPDD & VisA & DTD & SDD & \\
\midrule
\multirow{4}{*}{Commercial}
& GPT-4o~\citep{openai2024gpt4o} & / & 69.6 & 60.3 & 63.5 & 69.9 & 65.7 & 65.8 \\
& GPT-4.1-nano~\citep{verifiedOpenAI2025GPT41} & / & 74.7 & 61.7 & 60.5 & 78.3 & 50.0 & 65.0 \\
& GPT-4.1~\citep{verifiedOpenAI2025GPT41} & / & 81.9 & 66.7 & 69.1 & 90.1 & 79.9 & 77.5 \\
& Claude-Sonnet-4~\citep{anthropic2025claude4} & / & 67.6 & 65.9 & 63.5 & 88.4 & 81.7 & 73.4 \\
\midrule
\multirow{11}{*}{Open Source}
& LLaVA-OneVision-SI~\citep{li2024llavaonevision} & 0.5B & 50.0 & 50.0 & 50.0 & 54.3 & 50.0 & 50.9 \\
& Qwen2.5-VL-Instruct~\citep{Qwen2.5-VL} & 3B & 62.6 & 52.9 & 58.4 & 64.4 & 50.3 & 57.7 \\
& AnomalyR1~\citep{chao2025anomalyr1} & 3B & 69.4 & 56.0 & 59.8 & 61.0 & 57.6 & 60.8 \\
& InternVL-2.5~\citep{chen2024internvl} & 4B & 56.6 & 59.1 & 53.7 & 81.3 & 64.1 & 63.0 \\
& Qwen2.5-VL-Instruct~\citep{Qwen2.5-VL} & 7B & 66.0 & 56.0 & 58.4 & 59.2 & 67.4 & 61.4 \\
& Qwen3-VL-Instruct (rerun)~\citep{Qwen3-VL} & 8B & 67.0 & 50.0 & 46.8 & 70.2 & 50.0 & 56.8 \\
& AnomalyGPT~\citep{gu2024anomalygpt} & 7B & 46.6 & 54.2 & 57.3 & 64.1 & 49.5 & 54.3 \\
& Anomaly-OV~\citep{xu2025towards} & 7B & 74.3 & 70.3 & 74.3 & 90.7 & 88.7 & 79.7 \\
& LLaVA-1.5~\citep{liu2023llava15} & 13B & 61.4 & 61.4 & 67.2 & 75.9 & 50.0 & 63.2 \\
& LLaVA-1.6~\citep{liu2024llavanext} & 34B & 53.7 & 50.0 & 53.9 & 52.7 & 50.0 & 52.1 \\
& Qwen2.5-VL-Instruct~\citep{Qwen2.5-VL} & 72B & 77.4 & 64.7 & 68.2 & 81.9 & 69.0 & 72.2 \\
\midrule
Hybrid
& IndusAgent (8B policy + prior API) & 8B & \cellcolor{best2}83.6 & \cellcolor{best2}72.7 & \cellcolor{best2}76.8 & \cellcolor{best2}95.6 & \cellcolor{best2}88.9 & 83.5 \\
\bottomrule
\end{tabular}
}
\end{table}

Table~\ref{model_comparison} summarizes image-level balanced accuracy across five benchmarks covering industrial workpieces and surface textures. This range of datasets evaluates inspection performance across different product structures and defect appearances. The literature-reported baselines provide a broader performance context, while IndusAgent is identified as a hybrid system combining an 8B policy with an external visual-prior API. To examine this combination under a shared evaluation setting, we further compare the full system with direct-model baselines and an inference-time restriction on prior access.

\textbf{Common-protocol Comparisons.}
We evaluate the direct Qwen3-VL-8B backbone, direct Qwen3-VL-Max, IAD-R1, the full system, and the full system with the prior disabled at inference. These comparisons address complementary aspects of system performance: the direct backbone establishes the starting capability before industrial post-training, IAD-R1 provides a post-trained inspection baseline, and direct Max inference evaluates the external model when used as the standalone inspector. The prior-disabled variant retains the full trained checkpoint and measures its sensitivity to removing prior access at inference. All rows use the same images, preprocessing, binary task semantics, and metric definitions, with model-appropriate answer extraction described in Appendix~\ref{appx:baseline_parsing}.

Under the common protocol (Table~\ref{tab:controlled_system}), IndusAgent obtains 83.5\% mean balanced accuracy, exceeding direct Max inference at 79.4\% by 4.1 percentage points. Disabling the prior reduces performance to 79.7\%, a decrease of 3.8 points that reflects the trained policy's dependence on inference-time prior availability and the accompanying prompt change. The training objective is examined separately through the weight-matched ungated control. Across three training seeds, the full and ungated policies achieve five-benchmark means of $83.5\pm0.4$ and $80.5\pm0.5$, respectively (Table~\ref{tab:seed_robustness}). This difference supports the benefit of the complete accuracy-gated objective under the reported training setup, while the system-level comparisons establish the performance of the resulting policy and tool configuration.

\begin{table}[htb]
\centering
\caption{\textbf{Common-protocol system-level balanced accuracy (\%).} Max access indicates use of the external Qwen3-VL-Max component during inference. Prior disabling uses the full trained checkpoint with that tool removed from the prompt; it does not remove prior access from training.}
\label{tab:controlled_system}
{\renewcommand{\arraystretch}{1.08}
\setlength{\tabcolsep}{3.6pt}
\fontsize{8}{9.6}\selectfont
\begin{tabular}{lcc|ccccc|c}
\toprule
Method & Policy & Max access & MVTec & MPDD & VisA & DTD & SDD & Avg. \\
\midrule
Qwen3-VL-8B direct & 8B & No
& 67.0 & 50.0 & 46.8 & 70.2 & 50.0 & 56.8 \\
IAD-R1, rerun & 7B & No
& 78.4 & 68.1 & 70.2 & 90.9 & 82.0 & 77.9 \\
Qwen3-VL-Max direct & API & Yes
& 80.2 & 68.9 & 73.5 & 90.8 & 83.4 & 79.4 \\
\midrule
Prior disabled at inference & 8B & No
& 81.1 & 70.4 & 69.3 & 92.1 & 85.8 & 79.7 \\
\rowcolor{best2}
IndusAgent & 8B & Yes
& 83.6 & 72.7 & 76.8 & 95.6 & 88.9 & 83.5 \\
\bottomrule
\end{tabular}
}
\end{table}

The IAD-R1 rerun uses Qwen2.5-VL-7B. Prior disabling makes no Max call but retains the trained checkpoint and its prior exposure during training. Its 3.8-point gap therefore reflects inference-time tool availability and the accompanying prompt change.

\begin{figure}[htb]
    \centering
    \includegraphics[width=0.85\linewidth]{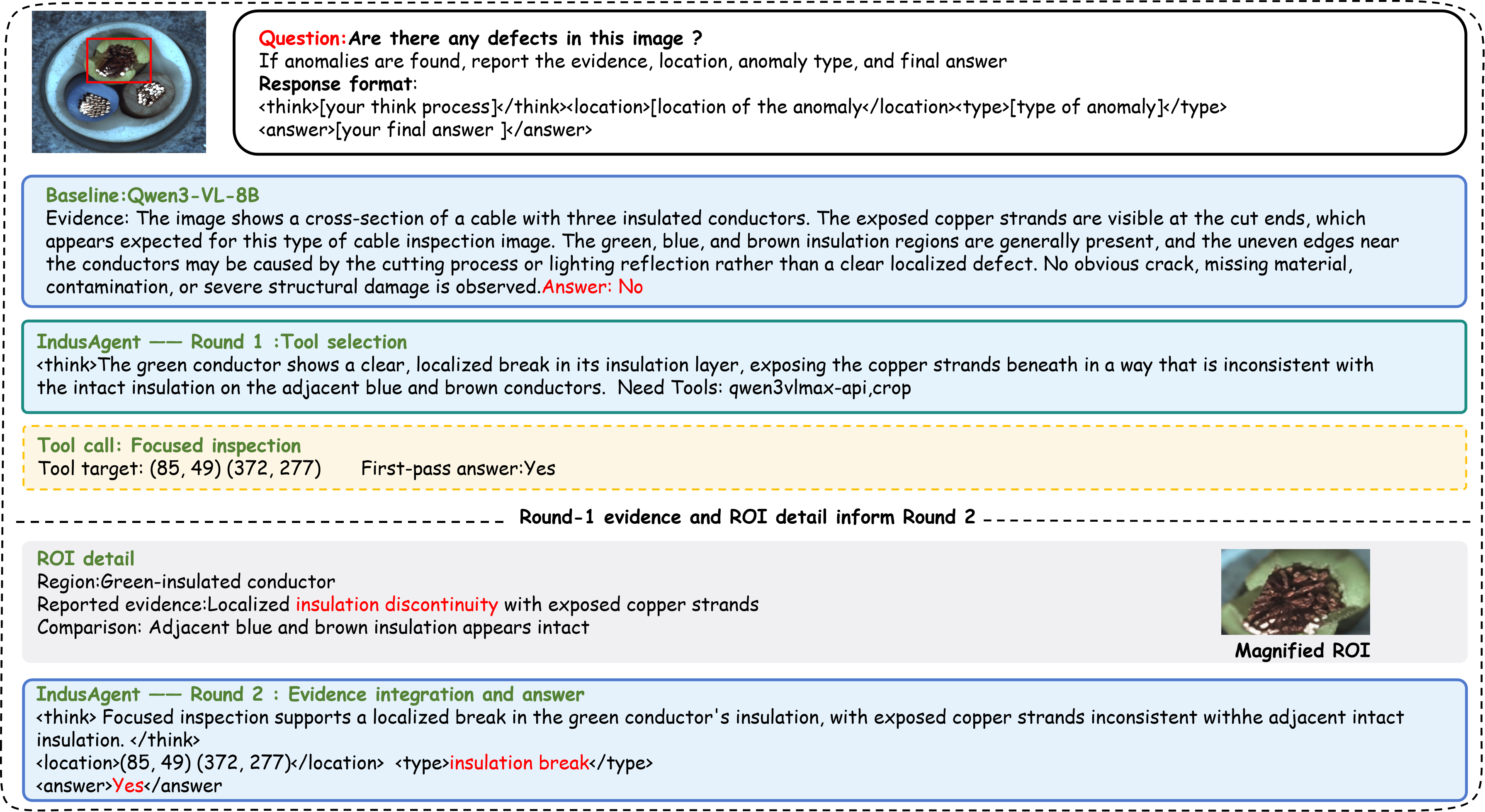}
    \caption{\textbf{Qualitative comparison on a cable image.}
    Direct Qwen3-VL-8B inference predicts no anomaly.
    IndusAgent flags a suspected insulation break, selects cropping
    and the external visual-prior tool, and integrates the available
    evidence into a final diagnosis with a defect box and type.}
    \label{fig:main_result_case}
\end{figure}

\noindent\textbf{Qualitative comparison.}
Figure~\ref{fig:main_result_case} contrasts the backbone's normal prediction with IndusAgent's suspected insulation break. The crop supports this suspicion through comparison with adjacent intact insulation. The second response retains the first-round anomaly judgment and reports the defect location and type.

\subsection{Classification and Grounding}

Table~\ref{tab:grounding_metrics} separates anomaly recall, normal-class specificity, and grounding quality. On MPDD, 95.4\% recall is accompanied by 50.0\% specificity: half of the normal images receive incorrect binary predictions under the evaluation protocol. This operating point would impose a substantial reinspection burden where false rejection is costly.

\begin{table}[htb]
\centering
\caption{\textbf{Classification and grounding of the full IndusAgent system (\%).}
Grounding metrics use all ground-truth anomalous images; missing or invalid fields receive zero.
Loc.@0.5 denotes the fraction with box IoU $\geq 0.5$.}
\label{tab:grounding_metrics}
{\renewcommand{\arraystretch}{1.08}
\setlength{\tabcolsep}{3.8pt}
\fontsize{8}{9.6}\selectfont
\begin{tabular}{l|cccc|ccc}
\toprule
Dataset & BA & AUROC & Recall & Specificity & Box mIoU & Loc.@0.5 & Type Acc. \\
\midrule
MVTec-AD & 83.6 & 92.1 & 85.5 & 81.7 & 48.7 & 46.2 & 72.8 \\
MPDD     & 72.7 & 88.5 & 95.4 & 50.0 & 41.5 & 38.6 & 68.9 \\
VisA     & 76.8 & 85.8 & 73.4 & 80.2 & 45.1 & 42.7 & 71.4 \\
DTD      & 95.6 & 95.8 & 94.1 & 97.1 & 59.8 & 61.5 & 80.7 \\
SDD      & 88.9 & 88.7 & 83.3 & 94.5 & 52.3 & 54.1 & 77.2 \\
\midrule
Macro Avg. & 83.5 & 90.2 & 86.3 & 80.7 & 49.5 & 48.6 & 74.2 \\
\bottomrule
\end{tabular}
}
\end{table}

\subsection{Ablation Studies}

We compare training-stage variants and test-time restrictions on three benchmarks. These interventions answer different questions and are specified below; additional results appear in Appendix~\ref{appx:Experimentapp}.

\begin{table}[htb]
\normalsize
\noindent\textbf{Training Stages and Inference Restrictions.}\par
\noindent
\begin{minipage}[t]{0.47\linewidth}
\vspace{0pt}\vspace{\abovecaptionskip}
Table~\ref{tab:module_ablation} compares SFT only with all tools, RL from the base model under the same reward and tool environment, and the full checkpoint with all six tool descriptions removed at inference. The last variant is evaluated without retraining, retaining two responses with empty call and observation lists.
On VisA, the full system achieves 76.8\% BA, versus 57.6\% for SFT only, 55.5\% without SFT, and 67.5\% with tools disabled. Even without tools, the policy outperforms SFT only; however, prompt changes prevent isolating the contribution of executed observations.
\end{minipage}\hfill
\begin{minipage}[t]{0.50\linewidth}
\vspace{0pt}
\centering
\caption{\textbf{Training-stage comparisons and test-time tool restrictions.}
Image-level balanced accuracy (\%) is reported on MVTec-AD, VisA, and DTD.
The last row disables all six tools at inference; the prior-disabled variant in Table~\ref{tab:controlled_system} removes only the visual-prior tool.}
\label{tab:module_ablation}

\fontsize{8}{9.6}\selectfont
\setlength{\tabcolsep}{3pt}
\renewcommand{\arraystretch}{1.08}

\begin{tabular}{@{}l|ccc@{}}
\toprule
\multicolumn{1}{c|}{Method} & MVTec & VisA & DTD \\
\midrule
\emph{(a)} Qwen3-VL-8B
& 67.0 & 46.8 & 70.2 \\
\emph{(b)} IndusAgent
& \cellcolor{best2}83.6
& \cellcolor{best2}76.8
& \cellcolor{best2}95.6 \\
\emph{(c)} SFT only (w/o RL)
& 72.3 & 57.6 & 74.1 \\
\emph{(d)} RL from base (w/o SFT)
& 69.5 & 55.5 & 72.8 \\
\emph{(e)} Tools disabled at inference
& \cellcolor{best}78.1
& \cellcolor{best}67.5
& \cellcolor{best}87.9 \\
\bottomrule
\end{tabular}
\end{minipage}
\end{table}
\FloatBarrier

\begin{table}[htb]
\centering
\caption{\textbf{Accuracy-gate and reward ablations (\%).}
Avg., grounding, and format metrics are macro-averages over MVTec-AD, VisA, and DTD.
Grounding follows Table~\ref{tab:grounding_metrics}; format failure measures two-round schema violations over all queries.
Base reward is $R_{\mathrm{acc}}+R_{\mathrm{format}}$.
Ungated removes only the accuracy gate; other variants remove the named term, including both bonus and call cost for tool.}
\label{tab:reward_gate}

\small
\setlength{\tabcolsep}{3.5pt}
\renewcommand{\arraystretch}{1.15}

\resizebox{\linewidth}{!}{%
\begin{tabular}{l|cccc|ccc|c}
\toprule
\multirow{2}{*}{RL reward}
& \multicolumn{4}{c|}{Balanced accuracy $\uparrow$}
& \multicolumn{3}{c|}{Grounding $\uparrow$}
& \multirow{2}{*}{\makecell{Format fail.\\$\downarrow$}} \\
\cmidrule(lr){2-5}\cmidrule(lr){6-8}
& MVTec & VisA & DTD & Avg. & Box mIoU & Loc.@0.5 & Type Acc. & \\
\midrule
Base reward & 76.0 & 64.9 & 79.1 & 73.3 & 40.8 & 37.3 & 62.9 & 3.4 \\
Ungated additive & 80.2 & 70.1 & 91.3 & 80.5 & 47.9 & 46.8 & 72.2 & 1.6 \\
\rowcolor{best2}
Full accuracy-gated & 83.6 & 76.8 & 95.6 & 85.3 & 51.2 & 50.1 & 75.0 & 1.2 \\
\midrule
Gated w/o tool utility & 81.9 & 71.5 & 92.8 & 82.1 & 48.6 & 48.3 & 73.9 & 1.1 \\
Gated w/o localization & 79.5 & 68.8 & 89.6 & 79.3 & 38.7 & 34.9 & 72.6 & 1.4 \\
Gated w/o type & 78.5 & 72.8 & 90.6 & 80.6 & 49.6 & 47.5 & 60.9 & 1.0 \\
Gated w/o format & 76.6 & 65.7 & 82.5 & 74.9 & 45.8 & 42.9 & 68.8 & 12.8 \\
\bottomrule
\end{tabular}%
}
\end{table}
On MVTec-AD, VisA, and DTD (Table~\ref{tab:reward_gate}), the full accuracy-gated objective improves mean BA from 80.5\% to 85.3\% over the weight-matched ungated control, with gains of 3.3, 3.3, and 2.8 points in Box mIoU, Loc.@0.5, and type accuracy, respectively. Removing localization or type rewards primarily degrades the corresponding grounding metrics, while removing tool utility reduces BA to 82.1\%. Removing the format reward increases format failures from 1.2\% to 12.8\% and lowers BA to 74.9\%. Because malformed binary answers count as incorrect, this BA decline may partly reflect reduced output compliance.

\section{Related Work}

\textbf{Industrial Anomaly Detection and Category Transfer.}
Reconstruction methods such as DRAEM and DiffusionAD model defect-free appearance or reconstruct normal structure~\citep{zavrtanik2021draem,verifiedZhang2023DiffusionAD}. Feature-based approaches such as PaDiM, PatchCore, and flow models provide strong inspection baselines when their required training data are available~\citep{defard2021padim,roth2022patchcore,yu2021fastflow,rudolph2022csflow}. Vision-language methods including WinCLIP, AnomalyGPT, and Anomaly-OV extend inspection through text-image alignment and multimodal reasoning~\citep{jeong2023winclip,gu2024anomalygpt,xu2025towards}. Our evaluation focuses on transfer from the stated source-category pool without paired normal images at inference; zero-shot inspection itself is not a new task introduced here.

\textbf{Post-training for Industrial Reasoning.}
SFT and reinforcement learning have been used for language and multimodal reasoning~\citep{deepseekr1,peng2025lmmr1}. IAD-R1 and AnomalyR1 apply post-training to industrial anomaly reasoning~\citep{li2026iadr1,chao2025anomalyr1}. IndusAgent builds on these training methods using actual tool-feedback trajectories and a final-correctness gate over auxiliary reward. The classification ablations assess this combination within the reported setup.

\textbf{Tool-augmented Visual Agents.}
Toolformer, AgentTuning, and multimodal tool-use frameworks study learning or orchestrating external operations~\citep{verifiedSchick2023Toolformer,zeng2024agenttuning,liu2023llavapluslearningusetools,ma2024taco,zhao2025pyvision}. AgentIAD combines tool-integrated trajectories with SFT and reinforcement learning, including a PZ-only mode without normal-exemplar retrieval~\citep{miao2025agentiad}. Its anomalous trajectory construction obtains inspection regions from ground-truth masks and supplies annotation information during reasoning generation. Indus-CoT instead uses teacher-selected calls and executed observations, with annotations consulted only to screen completed trajectories. IndusAgent couples this data construction with source--target product-category exclusion and a final-correctness gate over auxiliary rewards. Its external visual prior provides a query-conditioned opinion without retrieving a paired normal image. These distinctions concern data construction, evaluation scope, and optimization.

\section{Conclusion}
We presented IndusAgent for cross-category industrial anomaly detection without paired normal exemplars. Indus-CoT initializes inspection behavior with teacher-selected actions, executed feedback, and post-generation annotation screening. The accuracy-gated objective then couples auxiliary inspection credit to final diagnostic correctness. The hybrid system achieves 83.5\% mean balanced accuracy across five benchmarks, with the weight-matched three-seed comparison supporting a classification benefit from the gated objective. Test-time restrictions quantify sensitivity to tool availability, and full-system metrics report grounding performance. Improving normal-image specificity and resolving the contributions of individual reward terms remain priorities for more reliable inspection.

\subsection*{AI use statement}
Generative AI was used to synthesize the teacher inspection trajectories in Indus-CoT and to provide external visual-prior responses during tool-augmented inspection.AI tools also assisted with language editing, figure preparation, and manuscript assessment and revision.The authors are responsible for verifying the scientific claims, experimental results, citations, and final manuscript.

\subsection*{Reproducibility statement}
The appendix specifies the source categories, post-generation screening, two-round reward protocol, pixel-coordinate box convention, metric aggregation, test-time tool restrictions, and training-set hyperparameter selection. Hyperparameters were fixed before target evaluation and were not adjusted using target results. Three-seed results are provided in Table~\ref{tab:seed_robustness}. Exact launch configurations, canonical type mappings, and sample manifests are required to reproduce the experiments.

\bibliography{references}
\bibliographystyle{conference}

\appendix
\section{Detailed Reward Design}
\label{appx:reward_design}

Equation~\ref{eq:gated_reward} is implemented with $\alpha=0.8$, $\beta=0.6$, $\gamma=0.5$, a tool bonus of 0.3, a per-call coefficient of 0.1, and an independent format weight of 0.05.

\paragraph{Ungated control.}
The weight-matched additive objective used in the reward ablation is
\begin{equation}
R_{\mathrm{add}}=R_{\mathrm{acc}}+\alpha R_{\mathrm{loc}}+\beta R_{\mathrm{type}}+\gamma R_{\mathrm{tool}}+R_{\mathrm{format}}.
\end{equation}
All component definitions and coefficients are shared with the gated objective, including the format reward and the tool-call cost.

\paragraph{Binary correctness and auxiliary fields.}
$R_{\mathrm{acc}}=1$ only when a parsed final answer equals the source label. For ground-truth anomalous images, the box reward is continuous IoU after validating both boxes as finite and non-degenerate in the original-image pixel coordinate system. Type reward is exact agreement after deterministic canonicalization within the 29-label vocabulary in Appendix~\ref{appx:canonical_types}, with unknown or missing types receiving zero. Both auxiliary terms are zero for ground-truth normal images. Parse failures receive zero binary and spatial/type credit. This runtime reward is distinct from the IoU$\geq0.5$ criterion used to screen teacher trajectories.

\paragraph{Tool counts and success condition.}
$N_{\mathrm{call}}$ counts parsed calls admitted to the tool-execution round, including repeated calls, up to a maximum of four. Excess requests are ignored and do not enter this count. An admitted call incurs cost even if execution fails. $N_{\mathrm{ok}}$ counts successful observation records from these calls whose tool is one of the six tools in $\mathcal T$. The bonus in Eq.~\ref{eq:tool_reward} requires $N_{\mathrm{ok}}>0$ and a positive margin change. A successful status establishes execution validity, not correctness or diagnostic usefulness of the returned evidence.

\paragraph{Decision-margin change.}
For the final predicted label $\hat y\in\{\texttt{Yes},\texttt{No}\}$ and opposite label $\bar y$, define
\begin{equation}
m(\hat y\mid h)=\log p(\hat y\mid h)-\log p(\bar y\mid h),\qquad
\Delta m=m(\hat y\mid h_{\mathrm{post}})-m(\hat y\mid h_{\mathrm{pre}}).
\end{equation}
The trainer records raw Yes/No logit differences at the first response's \texttt{<pre\_answer>} position and the second response's \texttt{<answer>} position, and orients both toward the final label. The raw-logit difference equals the log-probability difference at that position. The scoring context includes the generated history available at each decision, and the second context includes the tool feedback and additional reasoning. A margin increase can therefore reflect extra deliberation or changed context as well as returned observations. Missing tagged decision positions yield a zero margin; a missing final binary answer yields zero oriented margins. The reward helper treats non-finite or unparseable margin inputs as zero gain. No verbalized confidence is used. Because the contexts and generated histories differ between rounds, $\Delta m$ is a confidence proxy rather than an isolated estimate of the feedback's causal effect.

\paragraph{Format reward.}
\begin{equation}
R_{\mathrm{format}}=0.05\,\mathbb I[\text{trajectory satisfies the two-round schema}].
\end{equation}
A compliant RL trajectory has exactly two non-empty \texttt{<think>} blocks, a valid preliminary Yes/No field, and a final answer. Tool-using trajectories have parsed calls and a JSON observation list with a matching record count. No-tool trajectories have empty call and observation lists and still contain both response segments. For an anomalous final answer, a valid defect box in original-image pixel coordinates and a non-empty type in the second response precede the final answer; for a normal final answer, these final auxiliary fields are absent. Invalid structures receive zero format reward, not a negative format penalty. Inference uses the same two-round schema specified in Appendix~\ref{appx:prompt}.
\paragraph{Reward ordering.}
Under the four-call execution limit, the smallest possible tool contribution is $\gamma R_{\mathrm{tool}}=-0.2$. Since localization, type, and format rewards are nonnegative, a correctly diagnosed trajectory receives at least $0.8$, whereas an incorrectly diagnosed trajectory receives at most the independent format reward of $0.05$. Thus, before groupwise normalization, every correctly diagnosed trajectory has a higher return than every incorrectly diagnosed trajectory. This ordering prioritizes diagnostic correctness while allowing auxiliary terms to distinguish trajectories with correct diagnoses.
\paragraph{Scope of the gate.}
For trajectories that satisfy the output schema, localization and type rewards are already zero on incorrect diagnoses: a false negative has no final box or type, and a truly normal image is ineligible for either reward. On this subset, $R_{\mathrm{acc}}R_{\mathrm{loc}}=R_{\mathrm{loc}}$ and $R_{\mathrm{acc}}R_{\mathrm{type}}=R_{\mathrm{type}}$. An additional effect on auxiliary-field credit is possible when an inconsistent trajectory predicts No on an anomalous image but still provides annotation-matching auxiliary fields. If the auxiliary parser accepts those fields, the additive objective can assign them credit, whereas the gated objective suppresses it.

The tool term also changes on incorrect trajectories: both its bonus and its call cost are removed, leaving at most the format reward. The gated--ungated comparison therefore evaluates the complete coupling of diagnostic correctness, auxiliary credit, and tool costs. The classification and grounding results in Table~\ref{tab:reward_gate}, together with the three-seed classification results in Table~\ref{tab:seed_robustness}, support the benefit of this combined objective. They do not isolate the contribution of each gated component or measure the frequency of reward exploitation during training.
\paragraph{Class-dependent reward range.}
Correct anomalous predictions can receive localization and type rewards that are absent on normal images. This difference in reward range is a design asymmetry, but does not by itself establish a preference for anomalous predictions: GRPO normalizes rewards among rollouts of the same image, whose ground-truth label is fixed. In particular, a false-positive decision on a normal image receives no gated auxiliary credit. Class-stratified reward and error diagnostics are needed to test whether the training dynamics contribute to the observed normal-image errors. An alternative with call cost outside the correctness gate would change the training objective and requires a separately trained comparison; it is not evaluated here.

\section{Additional Experimental Results}
\label{appx:Experimentapp}

\subsection{More Experimental Details}
\label{appx:exp_details}

\textbf{Category-Disjoint Training Protocol.}
The category audit compares Real-IAD object names with the five evaluation benchmarks, including semantic matches such as \texttt{pcb} with VisA's PCB categories and \texttt{transistor1} with MVTec-AD's transistor category. The four overlapping categories, \texttt{pcb}, \texttt{toothbrush}, \texttt{transistor1}, and \texttt{zipper}, were excluded before constructing the final training corpus. All 3,000 training trajectories come from the following 26 categories:
\begin{quote}\small\raggedright\ttfamily
 audiojack, bottle\_cap, button\_battery, end\_cap, eraser, fire\_hood, mint, mounts, phone\_battery, plastic\_nut, plastic\_plug, porcelain\_doll, regulator, rolled\_strip\_base, sim\_card\_set, switch, tape, terminalblock, toy, toy\_brick, u\_block, usb, usb\_adaptor, vcpill, wooden\_beads, woodstick.
\end{quote}
This is a product-category exclusion protocol; it does not demonstrate that all visual attributes or canonical defect types are unseen during training.

\textbf{Trajectory Generation and Curation.}
The teacher generates an inspection trajectory without access to ground-truth labels, locations, or defect types. Model-selected tool calls are executed, and actual feedback is supplied before the final diagnosis. Annotations are consulted only after generation to screen the completed trajectory. We retain candidates that satisfy the output format and agree with the binary annotation; anomalous candidates must additionally satisfy box IoU$\geq0.5$ and canonical defect-type agreement. Tool observations must correspond to executed calls. Normal candidates must not fabricate a final defect box or type.

Initially, 2,747 trajectories passed screening. The 253 unsuccessful samples were discarded. We generated candidates on other images from the same allowed source pool and selected 253 additional trajectories that passed the same criteria, completing a corpus of 3,000 accepted trajectories with 1,500 normal and 1,500 anomalous images. Failed trajectories were not repaired using ground-truth-conditioned text or included as placeholders.

\begin{table}[t]
\centering
\caption{\textbf{Reported corpus-size comparison.} Balanced accuracy (\%) for the 3,000-trajectory corpus and the 2,747-trajectory accepted subset. The completed corpus contains accepted replacements, not rejected or annotation-repaired trajectories.}
\label{tab:corpus_size}
{\renewcommand{\arraystretch}{1.08}
\setlength{\tabcolsep}{6pt}
\small
\resizebox{\linewidth}{!}{\begin{tabular}{lcccccc}
\toprule
SFT corpus & MVTec & MPDD & VisA & DTD & SDD & Avg. \\
\midrule
All 3,000 trajectories & 83.6 & 72.7 & 76.8 & 95.6 & 88.9 & 83.5 \\
2,747 accepted trajectories & 83.9 & 72.4 & 77.1 & 95.4 & 89.2 & 83.6 \\
\bottomrule
\end{tabular}}}
\end{table}

\textbf{Evaluation Metrics and Inference Details.}
For each dataset, balanced accuracy is computed from the pooled image-level confusion matrix, and F1 treats anomalous images as positive. Categories are not averaged separately within a dataset. The five-dataset average assigns equal weight to each dataset. Full-model localization and type metrics use all ground-truth anomalous images, including those with incorrect binary predictions; missing or malformed required fields receive zero credit. IoU$\geq0.5$ is the localization hit threshold. Training and test inference always generate two responses. Without tools, the intervening call and observation lists are empty. Evaluation uses only the second response's final diagnosis and auxiliary fields.

\subsection{Canonical Defect Vocabulary}
\label{appx:canonical_types}

We use a fixed vocabulary of 29 defect labels for post-generation trajectory screening, reinforcement-learning type rewards, and canonical type accuracy. Table~\ref{tab:canonical_types} lists the complete vocabulary. Predicted types are evaluated by exact agreement after deterministic normalization and dictionary matching. Missing, unrecognized, or non-matching types receive zero type credit. Normal predictions omit the final type field. Cross-category transfer is therefore evaluated over unseen product categories within a fixed defect-label space.

\begin{table}[t]
\centering
\caption{\textbf{Fixed canonical defect vocabulary.}
The 29 labels used for defect-type supervision and evaluation.
Each cell contains one complete label.}
\label{tab:canonical_types}
\small
\setlength{\tabcolsep}{12pt}
\begin{tabular}{@{}ll@{}}
\toprule
Canonical label & Canonical label \\
\midrule
dark stain contamination & surface abrasion \\
foreign object & frayed area \\
scratch & frayed bristles \\
crack & frayed wire \\
surface pit & frayed fiber \\
local dent & frayed traces \\
deformation & chipped damage \\
missing structural part & broken tip \\
missing material & broken area \\
missing tooth & misaligned pins \\
missing pin & short circuit \\
missing solder pad & solder protrusion \\
missing bristle cluster & surface defect \\
missing corner & insulation break \\
missing trace & \\
\bottomrule
\end{tabular}
\end{table}

\subsection{Baseline Prompting and Answer Parsing}
\label{appx:baseline_parsing}

The common binary task and metric definitions apply to the rerun models in Table~\ref{tab:controlled_system}, including direct Qwen3-VL-8B and Qwen3-VL-Max. IndusAgent additionally uses the tool-routing protocol. It does not imply that literature-only rows used the same prompt or parser. For each query image, the model is asked to determine whether the image contains an anomaly and to provide a final answer in a normalized binary form, where \texttt{Yes} denotes anomalous and \texttt{No} denotes normal. No paired normal reference image, category-specific exemplar, or dataset-specific prompt is provided during these reruns.

For IndusAgent, the final prediction is always extracted from the second response's structured \texttt{<answer>...</answer>} field, including when no tool is used. For IAD-R1, we apply a rule-based parser fixed before examining target labels. The parser first searches the final-answer span for explicit decisions such as ``yes'', ``anomalous'', ``defective'', ``abnormal'', ``no'', ``normal'', and ``defect-free''. When both positive and negative expressions occur, only the final stated conclusion is used. Responses that remain ambiguous or malformed are counted as incorrect rather than manually relabeled.

\subsection{More Hyperparameter Configurations}

\begin{table}[htbp]
\caption{\textbf{Reward-weight selection by separate post-training evaluation on the source training set.} These are training-set metrics, not held-out development results; target benchmarks were not used for selection.}
\centering
{\renewcommand{\arraystretch}{1.08}
\setlength{\tabcolsep}{6pt}
\small
\resizebox{\linewidth}{!}{\begin{tabular}{ccc|cccc}
\toprule
$\alpha$ (Loc) & $\beta$ (Type) & $\gamma$ (Tool) & \makecell{Train-set\\BA} & \makecell{Train-set\\mIoU} & \makecell{Train-set\\Type Acc.} & Calls/query \\
\midrule
0.0 & 0.0 & 0.0 & 78.9 & 37.8 & 62.4 & 0.66 \\
0.8 & 0.0 & 0.0 & 80.7 & 49.1 & 63.0 & 0.70 \\
0.8 & 0.6 & 0.0 & 81.5 & 50.3 & 70.8 & 0.73 \\
1.0 & 1.0 & 1.0 & 81.8 & 51.0 & 72.5 & 1.28 \\
0.8 & 0.6 & 0.5 & 82.6 & 50.8 & 72.1 & 0.91 \\
\bottomrule
\end{tabular}}
}
\label{tab:train_reward_selection}
\end{table}

Reward weights were selected by a separate evaluation pass on the same 3,000-image source training set after training each configuration. Balanced accuracy was the primary selection metric; localization, type accuracy, and call frequency were secondary diagnostics. There was no independent development set. The selected weights were frozen before evaluation on the five target benchmarks and were not adjusted using target results. Reusing training images for selection can give optimistic source-domain estimates.

\begin{table}[t]
\centering
\caption{\textbf{Three-seed results for the central comparisons.} Values are balanced accuracy averaged equally across the five target benchmarks. The full and ungated rows correspond to their respective trained policies. For each seed, prior disabling uses the corresponding full checkpoint with the prior tool description removed at inference; it is not a separately trained policy.}
\label{tab:seed_robustness}
{\renewcommand{\arraystretch}{1.08}
\setlength{\tabcolsep}{7pt}
\small
\resizebox{\linewidth}{!}{\begin{tabular}{lcccc}
\toprule
Variant & Seed 1 & Seed 2 & Seed 3 & Mean $\pm$ std. \\
\midrule
Ungated additive & 79.9 & 80.7 & 80.8 & $80.5\pm0.5$ \\
Prior disabled at inference & 79.4 & 79.8 & 79.9 & $79.7\pm0.3$ \\
Full accuracy-gated IndusAgent & 83.1 & 83.5 & 83.9 & $83.5\pm0.4$ \\
\bottomrule
\end{tabular}}}
\end{table}

  All experiments were conducted on a single compute node equipped with four NVIDIA A100 GPUs, each with 80GB memory. We used Qwen3-VL-8B-Instruct as the backbone model. The supervised
  fine-tuning stage was trained for one epoch on the Indus-CoT corpus and took approximately 21.6 minutes. The subsequent reinforcement learning stage was implemented with our
  accuracy-gated, fixed-two-round GRPO training pipeline and optimized for one epoch using four generations per prompt. This stage took approximately 23.4 hours. During RL training, we
  used a per-device batch size of 1, gradient accumulation steps of 2, a maximum prompt length of 4096, a maximum completion length of 512, bfloat16 precision, gradient checkpointing, and DeepSpeed ZeRO-3.

\subsection{More Experimental results}

\begin{table}[!ht]
  \centering
  \caption{\textbf{Image-level F1 (\%).} For rerun IAD-R1 and IndusAgent results, anomalous images are positive and counts are pooled over all images within each dataset. $\dagger$ denotes literature-transcribed contextual values.}
  \label{tab:f1_comparison}
  \setlength{\tabcolsep}{4.5pt} 
  \footnotesize 
  \begin{tabular}{lcccc}
    \toprule
    \multicolumn{1}{c}{\textbf{Model}} & \textbf{DTD} & \textbf{MPDD} & \textbf{MVTec} & \textbf{VisA} \\
    \midrule
    Qwen3-VL-8B & 84.8 & 66.2 & 77.8 & 58.3 \\
Claude-4-Sonnet & 88.5 & 69.8 & 81.2 & 64.5 \\
IAD-R1 (Qwen2.5-VL-7B)~\citep{li2026iadr1} & \cellcolor{best}91.1 & \cellcolor{best}75.2 & \cellcolor{best}85.7 & \cellcolor{best}70.8 \\
    \midrule
    IndusAgent (Ours) & \cellcolor{best2}96.1 & \cellcolor{best2}78.2 & \cellcolor{best2}89.8 & \cellcolor{best2}78.2 \\
    \bottomrule
  \end{tabular}
\end{table}
Table~\ref{tab:f1_comparison} reports F1 scores on four shared datasets. IndusAgent has the highest displayed F1 on each dataset under the reported protocols. The separate test-time disabling experiments characterize the trained policy's dependence on tool availability.

\subsection{Tool Usage Analysis}

\begin{table}[!ht]
  \centering
  \caption{\textbf{Tool-call composition and frequency during target inference.} Each tool percentage is its share of all counted tool calls on the corresponding dataset, with repeated calls counted separately. Shares sum to 100\% up to rounding. Average calls is the total number of counted calls divided by all query images, including those requiring no tool. Execution success measures valid-status returns rather than diagnostic usefulness.}
  \label{tab:tool_usage}
  \setlength{\tabcolsep}{4.5pt}
  \footnotesize
  \begin{tabular}{lccc}
    \toprule
    \textbf{Metric} & \textbf{MVTec} & \textbf{VisA} & \textbf{DTD} \\
    \midrule
    Crop (\%)    & 47.2 & 42.9 & 29.8 \\
    Prior (\%)   & 14.1 & 19.2 & 6.5 \\
    Enhance (\%) & 16.1 & 14.8 & 36.3 \\
    Measure (\%) & 7.4  & 5.9  & 2.2 \\
    Compare (\%) & 9.1  & 11.7 & 4.2 \\
    Texture (\%) & 6.1  & 5.5  & 20.9 \\
    \midrule
    Avg. calls & 1.32 & 1.30 & 0.94 \\
    Success rate (\%) & 99.1 & 98.7 & 99.4 \\
    \bottomrule
  \end{tabular}
\end{table}

Table~\ref{tab:tool_usage} reports tool-call composition and calls per query on three benchmarks. For dataset $d$, let $C_{d,t}$ be the number of counted calls to tool $t$ and $N_d$ the total number of query images. The tool share is $100C_{d,t}/\sum_{u\in\mathcal T}C_{d,u}$, while average calls is $\sum_{u\in\mathcal T}C_{d,u}/N_d$. Repeated calls contribute separately; images with no calls remain in $N_d$. Crop accounts for the largest call share on MVTec and VisA, and enhancement on DTD. Execution success is the fraction of counted calls returning a valid success status. Prior calls query the external Qwen3-VL-Max API.

\begin{table}[!ht]
\centering
\caption{\textbf{Measured end-to-end inference efficiency.} For IndusAgent, latency includes generation of both policy responses, plus tool execution and API waiting when invoked, under the measured hardware and network conditions. Latency is computed by averaging dataset-level mean per-image latencies with equal weights. API costs are reported in Chinese yuan (CNY) per query.}
\label{tab:inference_efficiency}
{\renewcommand{\arraystretch}{1.08}
\setlength{\tabcolsep}{6pt}
\small
\resizebox{\linewidth}{!}{\begin{tabular}{lccccc}
\toprule
Method & Tool calls/query & Max-API calls/query & Latency (s) & Output tokens & \makecell{API cost/query\\(CNY)} \\
\midrule
Qwen3-VL-8B direct   & 0 & 0 & 0.84 & 108 & 0 \\
Qwen3-VL-Max direct  & 1 & 1 & 2.31 & 142 & 0.0038 \\
Prior disabled at inference & 0.79 & 0 & 1.13 & 184 & 0 \\
IndusAgent           & 0.96 & 0.17 & 1.46 & 196 & 0.0006 \\
\bottomrule
\end{tabular}}}
\end{table}

Table~\ref{tab:inference_efficiency} reports the measured end-to-end cost of the resulting inference paths. IndusAgent has a mean latency of 1.46 seconds per image, compared with 0.84 seconds for direct Qwen3-VL-8B and 2.31 seconds for direct Qwen3-VL-Max. The 0.62-second increase over direct 8B inference is an end-to-end system difference encompassing additional policy generation and selected tool execution. The full system makes 0.17 Max-API calls per query and incurs a reported API cost of CNY~0.0006 per query under the measured service conditions.  Prior disabling reduces latency to 1.13 seconds while retaining a policy trained with prior access. The 0.33-second difference includes changes in policy generation and tool routing and cannot be attributed solely to API waiting. The available timing measurements do not separate policy generation, local tool execution, and remote waiting. Deployment therefore requires assessing the end-to-end inspection cost against the required cycle time and throughput under the intended hardware and service conditions.

\FloatBarrier
\subsection{Image-Level AUROC Analysis}

\begin{table}[!ht]
  \centering
  \caption{\textbf{Image-level AUROC (\%).} Published ZSAD baselines ($\dagger$; contextual) and IndusAgent on five IAD benchmarks. DTD denotes DTD-Synthetic.}
  \label{tab:image_level_auroc}
  \setlength{\tabcolsep}{2.5pt}
  \scriptsize
  \begin{tabular}{lcccccc}
    \toprule
    \multicolumn{1}{c}{\textbf{Method}}
    & \textbf{MVTec}
    & \textbf{MPDD}
    & \textbf{VisA}
    & \textbf{DTD}
    & \textbf{SDD}
    & \textbf{Avg.} \\
    \midrule
    CLIP~\citep{radford2021learning}
    & 74.1 & 54.3 & 66.4 & 71.6 & 65.7 & 66.4 \\

    WinCLIP~\citep{jeong2023winclip}
    & 91.8 & 63.6 & 78.1 & 93.2 & 84.3 & 82.2 \\

    VAND
    & 86.1 & 73.0 & 78.0 & 86.4 & 79.8 & 80.7 \\

    AnomalyCLIP~\citep{verifiedAnomalyCLIP2024}
    & 91.5 & 77.0 & 82.1 & 93.5 & 84.7 & 85.8 \\

    \midrule
    \textbf{IndusAgent}
& \textbf{92.1}
& \textbf{88.5}
& \textbf{85.8}
& \textbf{95.8}
& \textbf{88.7}
& \textbf{90.2}  \\
    \bottomrule
\end{tabular}
\end{table}

Table~\ref{tab:image_level_auroc} reports image-level AUROC using the normalized log-probability margin between the ``Yes'' and ``No'' decision tokens at the second response's final-answer position as the anomaly score. The five IndusAgent values yield a macro-average of 90.2\%. The baseline values are contextual because they were quoted rather than rerun under the same implementation. Different scalar scoring functions can be compared by AUROC under a matched evaluation protocol; differences in score construction alone do not invalidate that metric.

\subsection{More Ablation Studies}

\begin{table}[!ht]
\centering
\caption{\textbf{Individual tools disabled at inference.} Every row uses the same full trained checkpoint, removing only the indicated tool description from the inference prompt. These are availability interventions, without retraining.}
\label{table:ablation_individual_tools}
\resizebox{0.78\textwidth}{!}{
\footnotesize
\begin{tabular}{l|ccc}
\toprule
\multicolumn{1}{c|}{Method} & MVTec & VisA & DTD \\
\midrule
 IndusAgent & \cellcolor{best2}83.6 & \cellcolor{best2}76.8 & \cellcolor{best2}95.6 \\
\midrule
Crop disabled at inference & 79.4 & 68.6 & 91.2 \\
Prior disabled at inference & 81.1 & 69.3 & \cellcolor{best}92.1 \\
Enhance disabled at inference & 80.5 & 70.8 & 88.8 \\
Measure disabled at inference & \cellcolor{best}81.5 & 70.2 & 91.5 \\
Compare disabled & 80.9 & \cellcolor{best}74.1 & 91.8 \\
Texture disabled & 81.4 & 72.2 & 91.7 \\
\bottomrule
\end{tabular}
}
\end{table}

\textbf{Individual Tool Ablations.} Table~\ref{table:ablation_individual_tools} disables one tool at a time through the inference prompt at the same full checkpoint. Crop removal gives the largest VisA decrease, while enhancement removal gives the largest DTD decrease among the reported variants. Removing measurement or the Qwen3-VL-Max prior also reduces balanced accuracy on all three datasets. These observations are specific to the evaluated policy and benchmarks.

\begin{table}[!ht]
\centering
\caption{\textbf{GRPO group-size selection.} Source metrics are from separate post-training evaluation on the 3,000-image training set; step time is measured during training.}
\label{tab:train_group_size}
\resizebox{0.55\textwidth}{!}{
\footnotesize
\begin{tabular}{c|ccc}
\toprule
Group size & \makecell{Train-set\\BA} & \makecell{Train-set\\mIoU} & Step time \\
\midrule
2 & 80.9 & 48.6 & 8.7 \\
4 & 82.6 & 50.8 & 17.4 \\
6 & 82.4 & 51.2 & 25.8 \\
8 & 81.9 & 51.0 & 34.1 \\
\bottomrule
\end{tabular}
}
\end{table}

\textbf{Impact of GRPO Group Size.} Table~\ref{tab:train_group_size} compares group sizes using training-set evaluation and measured training cost. Group size 4 was selected and fixed before target-benchmark evaluation, without adjustment using target results. The source metrics are not independent generalization estimates.

\FloatBarrier

\section{Discussion on Novelty and Problem Setting}
\label{appx:discussion_novelty}

\paragraph{Problem scope.}
The study evaluates transfer from 26 Real-IAD source product categories to the stated target benchmarks without paired normal exemplars. Category-disjoint and reference-free evaluation are defining constraints of this study, rather than claims that zero-shot IAD is newly introduced. The fixed canonical type vocabulary does not test unconstrained naming or disjoint defect-type generalization. Unknown pretraining overlap is outside the category audit.

\paragraph{Training-data contribution.}
Indus-CoT links teacher-selected actions to actual executed tool feedback and a final diagnosis. The teacher is not given annotations during generation; annotations screen completed candidates for binary agreement, box IoU$\geq0.5$, and canonical type agreement. The 3,000 accepted trajectories initialize the policy for online optimization. Selection favors teacher-solvable examples and can limit coverage of difficult defects.

\paragraph{Reward contribution and evidence.}
The correctness gate conditions auxiliary rewards and call costs on the final binary diagnosis while retaining a separate format reward. The weight-matched additive comparison evaluates this complete coupling. Three-seed results support a classification benefit within the reported training setup. As detailed in Appendix~\ref{appx:reward_design}, the gate changes auxiliary-field credit on inconsistent outputs and tool-related returns on incorrect diagnoses;

\paragraph{Relation to prior agentic inspection.}
AgentIAD already combines tool-integrated trajectories, SFT, and reinforcement learning, including inspection without exemplar retrieval~\citep{miao2025agentiad}. Our study combines teacher-selected, executed-feedback trajectories with product-category exclusion and a correctness-gated objective. The visual-prior tool supplies an external opinion about the query and may identify suspicious deviations, although it is instructed not to return the final binary decision. Test-time disabling measures the trained policy's sensitivity to tool availability.

\section{Tool Library Specifications}
\label{appx:tool_library}
The tool library provides observations selected by the policy. Each tool-execution round admits at most four parsed calls across the six tool types, including repeated calls; requests beyond this limit are ignored. Successful execution does not guarantee that the returned evidence is correct or interpreted correctly.

\subsection{Dynamic Region Cropping Tool \texorpdfstring{($T_{\text{crop}}$)}{(T-crop)}}
The teacher or policy selects image coordinates, and the environment executes the requested crop and returns the local image. The same model-selected-coordinate mechanism is used in teacher generation, online RL, and target inference. Ground-truth defect coordinates are not supplied to the crop tool. A crop window specifies what to inspect; it is distinct from the final anomaly box evaluated against the annotation. Cropping exposes a local view without creating new optical information.

\subsection{Normalcy Prior Explanation Tool \texorpdfstring{($T_{\text{prior}}$)}{(T-prior)}}
The external Qwen3-VL-Max API receives the query image, the
inspection question, and a fixed instruction template with
a policy-specified focus supplied through the tool's
\texttt{query} argument. The instruction requests expected
defect-free geometry, texture, repeated structure, and
plausible manufacturing variation, while prohibiting a final
anomaly decision. Thus, the returned text is a
query-conditioned visual opinion. Ground-truth annotation
metadata and paired normal reference images are not supplied
to this API call.

\begin{promptbox}{Normalcy-Prior Instruction Template}
\begin{lstlisting}[style=promptstyle]
Act only as an industrial normalcy-prior tool. From the query image and general industrial knowledge, describe the expected defect-free geometry, texture, repeated structure, and plausible manufacturing variation. Do not make the final anomaly decision and do not assume a paired reference image. You receive no category, benchmark, path, label, mask, bounding box, or other annotation metadata.
Complete inspection question: {inspection_question}
Policy-requested focus: {query}
\end{lstlisting}
\end{promptbox}

\subsection{Low-Level Visual Enhancer \texorpdfstring{($T_{\text{enhance}}$)}{(T-enhance)}}
The tool returns deterministic transformations such as Canny edges and contrast-limited adaptive histogram equalization (CLAHE). These transformations may make contrast or boundaries easier to inspect, but can also amplify artifacts. The agent must interpret them alongside the query image.

\subsection{Geometric Verifier \texorpdfstring{($T_{\text{measure}}$)}{(T-measure)}}
The tool receives model-selected image coordinates and returns pixel distances and angular relations. It does not use camera calibration or report physical dimensions.

\subsection{Within-Image Comparison Tool \texorpdfstring{($T_{\text{compare}}$)}{(T-compare)}}
The tool compares a target region with other regions in the
same query image. It takes a normalized \texttt{target\_box}
and exactly one of \texttt{search\_box} or
\texttt{reference\_boxes}. Automatic search uses normalized
correlation, excludes candidates with excessive target
overlap, and applies spatial non-maximum suppression.
The tool supports edge and grayscale comparison, with up to
five references and three by default. Alignment is limited
to three pixels in each direction at the working resolution,
whose longest side is at most 384 pixels. It returns the
selected regions, matching and difference measurements, and
visual comparison artifacts. Comparison regions are not
assumed to be normal; the measurements describe differences
within the query image.

\subsection{Texture Analysis Tool \texorpdfstring{($T_{\text{texture}}$)}{(T-texture)}}
The tool analyzes periodic structure within a normalized
region of interest. It removes a fitted illumination plane,
applies a Hann window, and computes non-DC Fourier power
globally and in local windows. Periodicity quality is the
strongest eligible spectral-bin power divided by the total
eligible power. Local deviation combines differences in
period, orientation, and periodicity quality with weights
0.45, 0.45, and 0.10, respectively. Unreliable local windows
are assigned a deviation of 1, and overlapping window scores
are averaged to form the heatmap. Periods are reported in
original-image pixels and orientations in degrees modulo
180. The heatmap describes variation within the selected
region and is not a calibrated defect probability.
Regions with insufficient stable periodicity return an error.

\section{Prompt Protocols and Output Schemas}
\label{appx:prompt}
The following describes the information flow and structured fields for data generation, training, and evaluation. The schemas summarize the protocol; they are not presented as verbatim archived prompt transcripts. XML-style tags describe model output structure, while observations are inserted by the environment after tool execution.

\paragraph{Teacher generation and SFT.}
The teacher receives the query image, inspection instruction, and available tool descriptions. It inspects the image, requests tools if needed, receives actual returned observations, and generates its final diagnosis. Neither the generation instruction nor the tool context reveals the binary label, annotated location, or defect type. Those annotations are used only outside the teacher interaction, in post-generation screening. The accepted trajectories provide SFT targets.

\begin{promptbox}{Teacher Information Flow (Protocol Summary)}
\begin{lstlisting}[style=promptstyle]
INPUT: query image + inspection instruction + tool descriptions
TEACHER: visual inspection and optional tool requests
ENVIRONMENT: execute requested tools and return actual observations
TEACHER: final binary diagnosis; defect box and type when anomalous
SCREENING, AFTER GENERATION: format and binary-label agreement;
  for anomalous images, box IoU >= 0.5 and canonical-type agreement
RETAIN: accepted trajectory; otherwise discard the candidate
\end{lstlisting}
\end{promptbox}

\paragraph{Online RL.}
Each rollout has two generated response segments, regardless of whether the first requests a tool. Both segments contribute policy tokens to the optimization loss. The first response contains visual reasoning and a preliminary binary decision, followed by calls if requested. The environment inserts actual tool observations; the second response contains further reasoning and the final decision. No-tool trajectories emit an empty call list and receive an empty observation list before the second reasoning block and final decision. The following schemas use placeholders rather than specifying an unverified tool-argument JSON format.

\begin{promptbox}{Online RL Schema with Tools}
\begin{lstlisting}[style=promptstyle]
FIRST POLICY RESPONSE:
<think>[initial visual reasoning]</think>
<pre_answer>[Yes or No]</pre_answer>
<call_tool>[parsed tool calls with model-selected arguments]</call_tool>

ENVIRONMENT INSERTION:
<observation>[JSON list of actual execution records]</observation>

SECOND POLICY RESPONSE:
<think>[reasoning with available feedback]</think>
[If Yes: <location>[pixel box]</location><type>[canonical label]</type>]
<answer>[Yes or No]</answer>
\end{lstlisting}
\end{promptbox}

\begin{promptbox}{Online RL Schema without Tools}
\begin{lstlisting}[style=promptstyle]
FIRST POLICY RESPONSE:
<think>[initial visual reasoning]</think>
<pre_answer>[Yes or No]</pre_answer>
<call_tool>[]</call_tool>

ENVIRONMENT INSERTION:
<observation>[]</observation>

SECOND POLICY RESPONSE:
<think>[final visual reasoning]</think>
[If Yes: <location>[pixel box]</location><type>[canonical label]</type>]
<answer>[Yes or No]</answer>
\end{lstlisting}
\end{promptbox}

\paragraph{Target inference.}
Target inference follows the same fixed two-round protocol as online RL. The first response provides a preliminary diagnosis and selects zero or more tool calls. If no tool is selected, the policy outputs an empty call list, and the environment inserts an empty observation list. The second response is generated in all cases and provides the final diagnosis, together with the defect location and canonical type when predicting an anomaly.

\begin{promptbox}{Two-round Protocol without Tools}
\begin{lstlisting}[style=promptstyle]
FIRST POLICY RESPONSE:
<think>[initial visual reasoning]</think>
<pre_answer>[Yes or No]</pre_answer>
<call_tool>[]</call_tool>

ENVIRONMENT INSERTION:
<observation>[]</observation>

SECOND POLICY RESPONSE:
<think>[final visual reasoning]</think>
[If Yes: <location>[pixel box]</location><type>[canonical label]</type>]
<answer>[Yes or No]</answer>
\end{lstlisting}
\end{promptbox}

\paragraph{Tool availability interventions.}
Individual-tool and all-tool disabling remove the corresponding descriptions from the inference prompt at the same full trained checkpoint. The two-round structure is retained; with all tools disabled, the call and observation lists are empty. Other evaluation settings remain fixed. The no-RL row uses the SFT checkpoint with the full tool library, while the no-SFT row is trained by applying the same RL reward and executed tool environment directly to the base model.

\paragraph{Direct-model evaluation.}
Direct baselines inspect the query without paired normal references and produce a binary anomaly decision. The following supplied template documents the structured output convention; the exact model-specific wrappers and decoding configurations are required for a complete reproduction.
\begin{promptbox}{Direct Inspection Output Template}
\begin{lstlisting}[style=promptstyle]
You are an expert industrial quality control inspector.
Inspect the given image carefully for localized anomalies.
Output ONLY this format:
<think>short visual evidence</think>
<location>[x1,y1,x2,y2] in original-image pixels, or none</location>
<type>canonical defect label, or none</type>
<answer>Yes or No</answer>
\end{lstlisting}
\end{promptbox}

\newpage

\section{Detailed Related Work}
\label{appx:detailed_related_work}

This section expands the discussion in the main paper by connecting industrial anomaly detection with research on multimodal reasoning, evidence acquisition, spatial perception, and domain transfer.
We first discuss methods that directly address industrial inspection, then examine related developments in tool use, post-training, structured visual understanding, and domain-specific perception.
The broader connections concern how a model acquires and verifies information before making a decision; they do not imply that methods developed for other tasks have been evaluated as industrial anomaly detectors.

\paragraph{Industrial anomaly detection and category transfer.}
Industrial inspection methods differ in how they represent normal appearance and identify deviations from it.
DRAEM learns anomaly representations through a reconstruction-based discriminative framework~\citep{zavrtanik2021draem}.
DiffusionAD uses diffusion-based reconstruction for anomaly detection~\citep{verifiedZhang2023DiffusionAD}.
Feature-distribution methods offer a complementary approach: PaDiM models patch-level feature distributions, while PatchCore stores representative normal patch features for nearest-neighbor comparison~\citep{defard2021padim,roth2022patchcore}.
FastFlow and CS-Flow instead use normalizing flows to model visual feature distributions~\citep{yu2021fastflow,rudolph2022csflow}.
Vision-language alignment introduces semantic descriptions of normal and anomalous appearance, as exemplified by WinCLIP and AnomalyCLIP~\citep{jeong2023winclip,verifiedAnomalyCLIP2024}.
AnomalyGPT extends inspection to multimodal dialogue and localized anomaly understanding~\citep{gu2024anomalygpt}.
Anomaly-OV further studies open-vocabulary industrial anomaly detection with a multimodal language model~\citep{xu2025towards}.
Industrial post-training then adapts the reasoning process itself, with IAD-R1 and AnomalyR1 investigating reinforcement learning for anomaly reasoning~\citep{li2026iadr1,chao2025anomalyr1}.
AgentIAD incorporates tool-integrated trajectories into supervised fine-tuning and reinforcement learning, and supports inspection with or without normal-exemplar retrieval~\citep{miao2025agentiad}.
Within this line of work, IndusAgent studies category-disjoint transfer with teacher-selected inspection actions, executed tool observations, and a final-correctness gate over auxiliary rewards.
Its reference-free inference protocol excludes paired normal exemplars while retaining access to an external visual-prior API.

\paragraph{Tool-augmented visual reasoning and evidence acquisition.}
The motivation for tool use is that additional reasoning can require observations that are absent from the model's initial view.
Toolformer studies learning when and how to invoke external APIs from language-model supervision~\citep{verifiedSchick2023Toolformer}.
LLaVA-Plus extends tool-use learning to multimodal assistants equipped with external visual capabilities~\citep{liu2023llavapluslearningusetools}.
TACO trains multimodal models to select and use tools through action-oriented reasoning traces~\citep{ma2024taco}.
PyVision develops visual reasoning through executable programs that manipulate and inspect visual inputs~\citep{zhao2025pyvision}.
For open-vocabulary action recognition, Video-STAR combines contextual sub-motion decomposition with tool-augmented reinforcement learning~\citep{yuan2025video}.
FactGuard applies agentic supervised fine-tuning and decision-aware reinforcement learning to acquire evidence for video misinformation verification~\citep{li2026factguard}.
ImagineAgent introduces generative imagination into tool-assisted open-vocabulary human-object interaction comprehension~\citep{yuan2026if}.
At the workflow level, AutoDesign uses rollout feedback to optimize the harness of a long-horizon design agent~\citep{luo2026autodesign}.
These studies connect tool selection with the information needed to resolve a task, although their action spaces and verification targets differ.
IndusAgent specializes this connection to industrial inspection, where cropping, enhancement, measurement, comparison, texture analysis, and external visual opinions support a structured defect diagnosis.

\paragraph{Reinforcement learning, reflection, and grounded reward signals.}
Post-training provides a means of optimizing the reasoning policy beyond imitating demonstrations.
DeepSeek-R1 investigates reinforcement learning for reasoning, while GRPO provides group-relative policy optimization without a separate value model~\citep{deepseekr1,shao2024deepseekmath}.
ReflectRL extracts learning signals from failed expert trajectories through reflective reasoning and transfers this behavior back to direct reasoning~\citep{bi2026reflectrllearninggoldennegative}.
EchoRL recovers supervision from successful rollouts whose group-relative advantages have degenerated, using selected trajectory segments as auxiliary training signals~\citep{bi2026echorl}.
The Geometry of Reasoning studies self-evaluation through the layerwise evolution of reasoning trajectories~\citep{bi2026the}.
Rubric-guided reasoning supplies fine-grained rewards and offline guidance to broaden exploration during multi-domain GRPO training~\citep{bi2025reward}.
PromptCD controls behavior at test time by contrasting positive and negative guiding prompts, including their effects on visual attention in vision-language models~\citep{bi2026promptcd}.
For video reward modeling, CaC combines temporal concentration, spatial grounding, and two-turn GRPO with localization rewards~\citep{wang2026cac}.
These approaches address complementary aspects of training signals, self-evaluation, and inference-time control.
IndusAgent concentrates on the relationship between final diagnostic correctness and auxiliary inspection credit: localization, defect-type, and tool-related rewards are gated by the binary diagnosis, while format supervision remains independent.
Its reported ablations assess this reward coupling within the stated industrial training protocol.

\paragraph{Multimodal architectures and training-data quality.}
The information available to a reasoning policy depends on both its representation and the supervision used to adapt it.
mc-BEiT replaces single-token targets in masked image modeling with refined multi-choice supervision, providing a different approach to learning discrete visual representations~\citep{li2022mc}.
LongCat-Next represents text, vision, and audio through discrete tokens under a shared autoregressive modeling framework~\citep{team2026longcat}.
Language-Pretraining-Induced Bias investigates adapting language-pretrained parameters to visual tasks through random-label bridge training, including partial adaptation that retains useful pretrained layers~\citep{luo2026language}.
For efficient multimodal computation, $\gamma$-MoD uses attention-map rank to select layers for mixture-of-depth adaptation, together with shared vision-language routing and masked routing learning~\citep{luo2025gamma}.
DenseFusion-1M integrates vision experts into a captioning engine to construct dense image descriptions for multimodal perception training~\citep{li2024densefusion}.
Data selection offers another route to adaptation: PRISM studies training-free selection of multimodal instruction data by accounting for intrinsic visual semantics~\citep{Bi2025PRISMSI}.
RefineX distills expert-guided text refinement into editing programs for large-scale pre-training corpus improvement~\citep{bi2025refinex}.
LLMSurgeon studies post-hoc estimation of an LLM's domain-level pretraining mixture from generated text, using a calibrated confusion matrix and a constrained inverse problem under a label-shift assumption~\citep{luo2026llmsurgeon}.
Together, these directions show how representation design and data quality shape the inputs from which a model learns to reason.
Indus-CoT addresses a more specific supervision need by pairing teacher-selected inspection actions with actual tool feedback and screening completed diagnoses against annotations.
It adapts inspection behavior on a fixed backbone; the category audit concerns industrial post-training and does not establish the composition of the backbone's pretraining corpus.

\paragraph{Model transferability and out-of-distribution generalization.}
Cross-category inspection requires considering both the suitability of a pretrained representation and its response to changes in visual appearance.
Potential-energy-based transferability estimation models the evolution of pretrained features to assess their suitability for downstream tasks~\citep{li2023exploring}.
Uncertainty Modeling for Out-of-Distribution Generalization represents feature statistics as uncertain quantities and perturbs them during training to improve robustness to potential domain shifts~\citep{li2022uncertainty}.
The former addresses model selection before adaptation, while the latter modifies feature-level training to accommodate distributional variation.
IndusAgent evaluates transfer through explicit source--target product-category exclusion and learns an inspection policy that can request local evidence from each query.
Its experiments concern this training and inference protocol rather than a transferability estimator or an uncertainty-based feature augmentation scheme.

\paragraph{Human-object interaction, spatial reasoning, and embodied perception.}
Structured visual understanding requires connecting semantic hypotheses with the entities and regions that support them.
Synergistic prompting learning combines visual and textual prompting for human-object interaction detection~\citep{luo2025synergistic}.
InstructHOI uses context-aware instructions to guide multimodal reasoning about ambiguous interactions~\citep{luo2026instructhoi}.
Human-centric open-future task discovery extends human-centered understanding to identifying useful tasks across possible futures through structured search~\citep{song2026human}.
SIRI-Bench evaluates structural spatial intelligence using problems that require both spatial comprehension and reasoning in three-dimensional scenes~\citep{song2025siri}.
In robotic manipulation, vision-to-geometry mapping conditions action generation on geometric representations rather than relying primarily on language or video backbones~\citep{song2026robotic}.
Pelican-Unify~1.0 combines understanding, reasoning, imagination, and action within a unified embodied model~\citep{zhang2026pelican}.
Pelican-VLA~0.5 studies how a compact perception-to-action bottleneck supports attention to manipulation-relevant objects and contact regions~\citep{ding2026pelican}.
These tasks share a need to associate an interpretation with spatially relevant evidence, but they predict interactions or actions rather than product defects.
For industrial inspection, the corresponding challenge is to verify whether a suspicious local structure supports an anomalous diagnosis within the fixed defect vocabulary.

\paragraph{Stereo reconstruction, geometric priors, and boundary-sensitive perception.}
Research on stereo and three-dimensional reconstruction provides related perspectives on preserving local geometry under weak texture, occlusion, and uncertain correspondence.
EHSS investigates efficient stereo matching through a hybrid-supervised symmetric network~\citep{zhang2023ehss}.
MonoMVSNet introduces monocular priors into multi-view stereo, whereas MVSMamba explores state-space modeling for multi-view reconstruction~\citep{monomvsnet,mvsmamba}.
MSP-MVS uses multi-granularity segmentation priors to guide stereo estimation~\citep{yuan2025msp}.
DVP-MVS combines depth-edge and visibility priors to improve geometric matching~\citep{yuan2025dvp}.
DVP-MVS++ extends this direction with depth, normal, and edge information together with harmonized visibility priors~\citep{yuan2025dvpplusplus}.
Segmentation can also guide the matching support itself: SD-MVS studies segmentation-driven deformation with spherical refinement, while TSAR-MVS addresses textureless regions through segmentation and correlative refinement~\citep{yuan2024sd,yuan2024tsar}.
SED-MVS further combines edge-aligned deformation, depth restoration, and occlusion constraints~\citep{yuan2025sed}.
For Gaussian-based scene representations, GradiSeg uses gradient-guided refinement to improve semantic segmentation near three-dimensional boundaries~\citep{li2024gradiseg}.
These methods reconstruct or segment geometry from their respective observations; IndusAgent currently performs inspection on a single query image.
The connection lies in the importance of boundary and structural evidence, which the inspection policy can examine through local cropping, enhancement, and geometric measurement without estimating a full three-dimensional reconstruction.

\paragraph{Scene generation, geometric verification, and adaptive visual computation.}
Generative modeling can encode structural priors, but its outputs need to be interpreted according to the task that produced them.
Layout2Scene conditions three-dimensional scene generation on semantic layouts and geometry and appearance diffusion priors~\citep{chen2025layout2scene}.
Graph2Scene organizes indoor scene generation with interaction-aware scene graphs~\citep{chen2025graph2scene}.
STDR decouples spatial and temporal factors in Gaussian-based dynamic scene representations to improve reconstruction consistency~\citep{li2025stdr}.
Geometry-guided reinforcement learning for scene editing, instantiated as RL3DEdit, uses geometric verification signals to promote consistency across edited views~\citep{wang2026geometry}.
From Editor to Dense Geometry Estimator adapts structural priors from an image-editing model to depth and normal prediction~\citep{wang2026edit}.
Adaptive test-time scaling for image editing allocates computation according to edit difficulty and uses edit-specific verification and stopping criteria~\citep{qu2026scale}.
These studies connect prior knowledge, structural constraints, and the allocation of visual computation.
IndusAgent's present toolkit acquires image-derived observations and an external textual opinion rather than synthesizing missing views or reconstructing a scene.
Its two-round protocol uses a fixed number of response rounds, with optional tool calls inside that protocol.

\paragraph{Vision-language-action reasoning and autonomous-driving transfer.}
Autonomous-driving research studies how visual representations and reasoning can support decisions under changing environments.
The survey of pure vision-language-action models organizes approaches that connect perception, language, and action generation~\citep{zhang2025pure}.
Reasoning-VLA combines reasoning-enhanced visual-language features with parallel action-trajectory generation and supervised and reinforcement learning fine-tuning~\citep{zhangreasoning}.
AutoDrive-R$^{2}$ uses chain-of-thought supervision, self-reflection, and physics-grounded reinforcement learning for trajectory planning~\citep{yuan2025autodrive}.
CoC-VLA studies causal visual-language-action reasoning together with adversarial transfer from simulated to real-world driving~\citep{NEURIPS2025_66d09284}.
AT-Drive also investigates adversarial transfer between simulation-trained and real-world autonomous-driving models~\citep{zhang2025drive}.
ADDI explores distinct experts and implicit interactions in a simplified end-to-end driving model~\citep{zhang2025addi}.
At the perception level, MapExpert studies efficient online high-definition map construction through sparse map-element experts~\citep{zhang2025mapexpert}.
This literature provides related examples of structured decision making and transfer, while the observations, actions, and supervision differ from industrial anomaly detection.
IndusAgent's transfer setting excludes target product categories from industrial post-training and evaluates a diagnostic decision rather than a driving trajectory.

\paragraph{Cross-modal retrieval and robust multi-view representation learning.}
Combining visual observations with textual requirements raises questions about which evidence should dominate the resulting representation.
ConeSep addresses noisy composition in image retrieval through a cone-based noise-unlearning network~\citep{ConeSep}.
TEMA anchors the reference image while mapping multiple textual modifications to relevant entities for composed image retrieval~\citep{TEMA}.
ReTrack studies composed video retrieval through evidence-driven dual-stream directional anchor calibration~\citep{ReTrack}.
For multi-view data, SEC-LSRM combines sampling-enhanced contrastive learning with long- and short-range information mining in remote-sensing clustering~\citep{SEC-LSRM}.
Prototype-driven multi-view attribute-missing graph clustering uses structural information across views to learn with incomplete attributes~\citep{PAGC}.
These methods concern retrieval or unsupervised representation learning, while industrial diagnosis must integrate local images, measurements, and textual observations into a decision about the query.
The related issue is maintaining the correspondence between each piece of evidence and the hypothesis it is intended to test.
IndusAgent does not retrieve paired normal images or learn a multi-view clustering model at inference.

\paragraph{Domain-specific visual interpretation and prompt adaptation.}
Applications in remote sensing and biomedical analysis illustrate how task structure changes the requirements for visual reasoning.
Decoupled prompt learning for remote-sensing image change captioning connects representations of visual change with a pre-trained language model for description generation~\citep{liu2023decoupling}.
The survey of remote-sensing spatiotemporal vision-language models reviews the integration of spatial, temporal, and linguistic information in this domain~\citep{liu2025remoteVLM}.
M$^{2}$Reg investigates unsupervised multi-scale registration for multimodal microscopy~\citep{wu2026m}.
EMPOWER optimizes medical prompts through an evolutionary framework with domain-specific evaluation and reinforcement learning~\citep{chen2026empower}.
These applications require domain-relevant interpretation or alignment, although the observations and error criteria differ from those of manufacturing inspection.
IndusAgent addresses this requirement through industrial trajectory supervision, a fixed diagnostic output schema, and a query-conditioned visual-prior tool.

Across these research areas, three recurring issues connect most directly to our setting: selecting observations that resolve ambiguity, grounding interpretations in local structure, and assigning training credit according to task-relevant correctness.
IndusAgent brings these issues together within the stated category-disjoint industrial inspection protocol, using executed-feedback trajectories and an accuracy-gated objective to train its two-round inspection policy.

 \section{More Case Studies}
  \label{appx:case_study}

\begin{figure} [h]
  \centering
  \includegraphics[width=\linewidth]{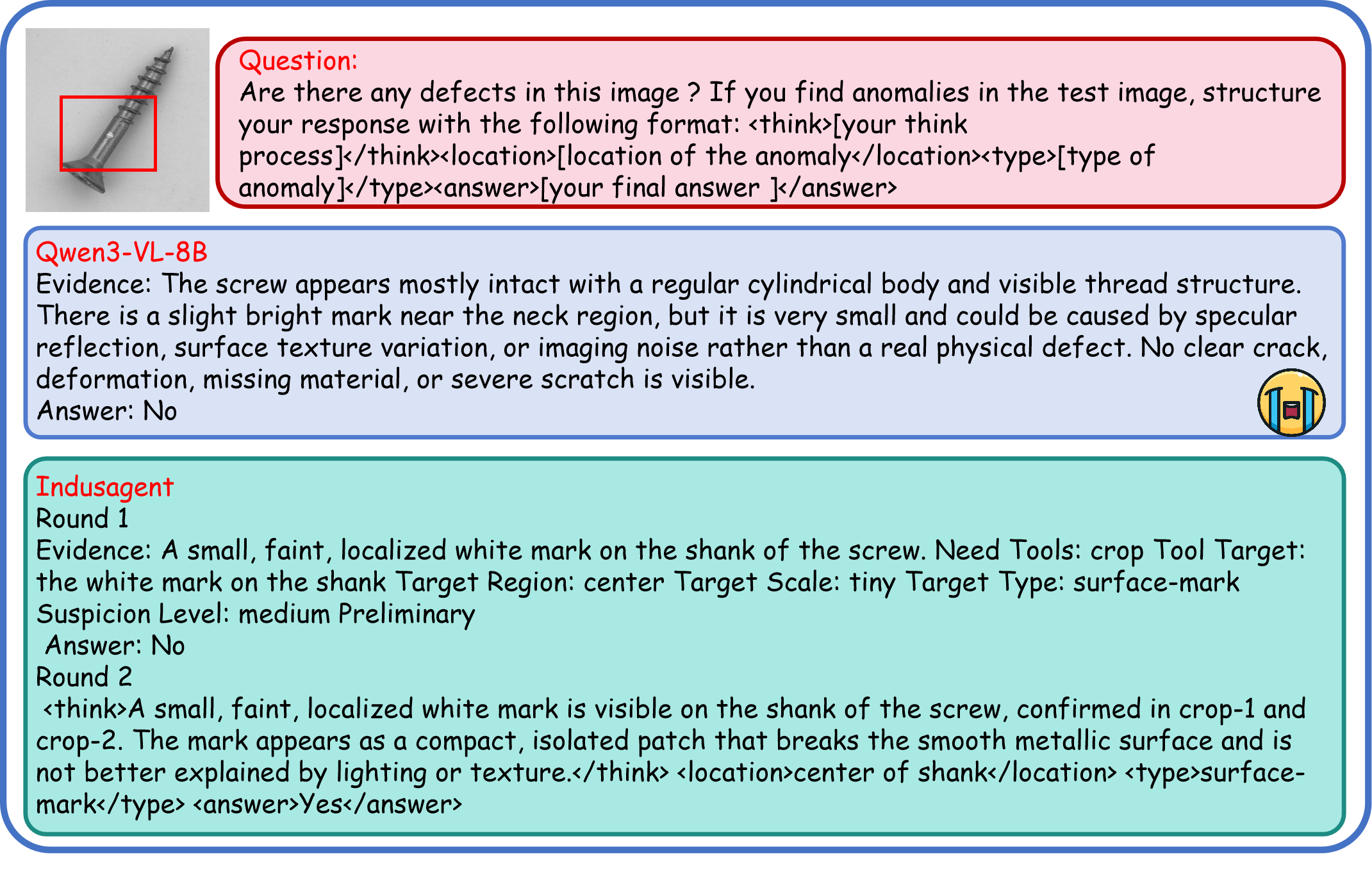}
   \caption{
    Case Study between Qwen3-VL-8B and our method.
    }
  \label{fig:case3}
\end{figure}

\begin{figure} [h]
  \centering
  \includegraphics[width=\linewidth]{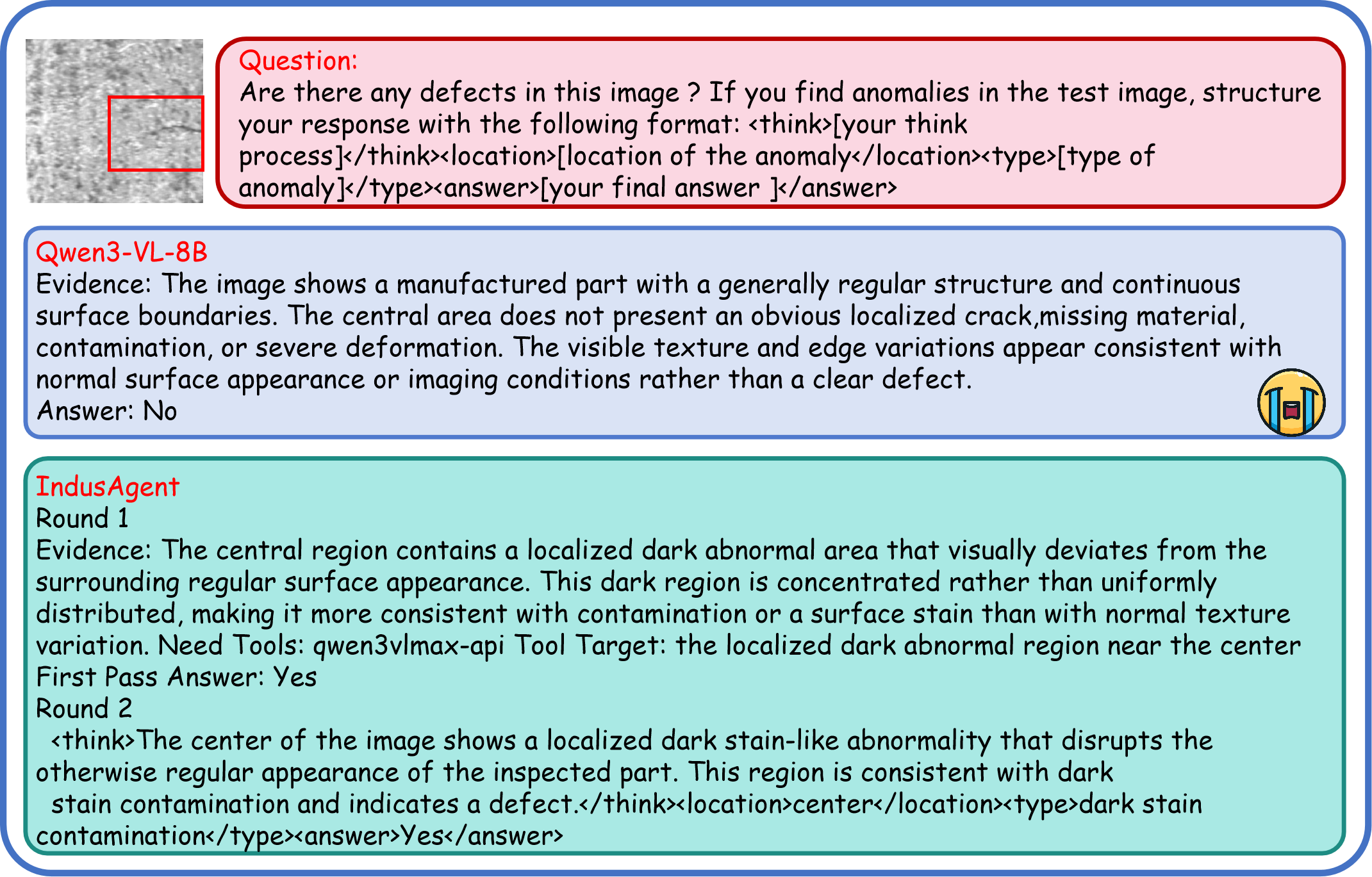}
   \caption{
    Case Study between Qwen3-VL-8B and our method.
    }
  \label{fig:case2}
\end{figure}

\section{Limitations}
\label{sec:limitations}

The evaluation uses public benchmarks rather than a production-line deployment. Bounding boxes provide coarser localization than masks, and a fixed canonical vocabulary does not evaluate unconstrained defect naming. The category audit concerns industrial post-training; unknown backbone and API pretraining overlap cannot be excluded. Teacher-success filtering can underrepresent difficult images, and the corpus may inherit teacher and prompt biases.

Reward weights and group size were selected on the source training set without an independent development set and were frozen before target evaluation; the source selection metrics may therefore be optimistic. The correctness gate removes call costs on incorrect trajectories and does not guarantee minimal tool use. Reward ablations in Table~\ref{tab:reward_gate} report classification, grounding, and format-compliance metrics on three benchmarks, but do not separate the effects of gating localization and type rewards from those of gating tool bonuses and call costs. Tool-usage and individual-tool availability analyses cover all six tools on three of the five benchmarks. Prompt-based disabling jointly changes available feedback and input context. The MPDD specificity of 50.0\% remains an unresolved normal-image error limitation.

The full system depends on an external Qwen3-VL-Max visual-prior API. The reported inference-time prior restriction retains prior exposure during training and does not establish the performance of a policy trained entirely without external visual opinions. Latency and cost describe the measured service and network conditions. Misleading tool outputs can affect diagnosis. Literature-transcribed baseline results provide context, and there is no matched AgentIAD rerun.

\end{document}

%% file: math_commands.tex
\usepackage{amsmath,amsfonts,bm}

\def\eqref#1{equation~\ref{#1}}

\def\1{\bm{1}}

\DeclareMathAlphabet{\mathsfit}{\encodingdefault}{\sfdefault}{m}{sl}
\SetMathAlphabet{\mathsfit}{bold}{\encodingdefault}{\sfdefault}{bx}{n}

